\PassOptionsToPackage{dvipsnames,table}{xcolor}
\documentclass{article} 
\usepackage{iclr2025_conference,times}

\usepackage{amsmath,amsfonts,bm}

\def\eqref#1{equation~\ref{#1}}

\def\1{\bm{1}}

\DeclareMathAlphabet{\mathsfit}{\encodingdefault}{\sfdefault}{m}{sl}
\SetMathAlphabet{\mathsfit}{bold}{\encodingdefault}{\sfdefault}{bx}{n}

\usepackage{hyperref}
\usepackage{url}
\usepackage{comment}
\usepackage{graphicx}
\usepackage{pifont}
\usepackage{booktabs}
\usepackage{listings}
\graphicspath{{./Images/}}
\usepackage{colortbl}
\usepackage{multirow}
\usepackage{tabularx}
\usepackage{adjustbox}
\usepackage{subcaption}
\usepackage{amssymb}
\usepackage{diagbox}
\usepackage{amsfonts}       
\usepackage{nicefrac}       
\usepackage{microtype}      
\usepackage{varwidth}
\usepackage{xcolor}  
\usepackage{indentfirst}
\lstdefinelanguage{yaml}{
    morecomment=[l]{\#},           
    morestring=[b]",               
    sensitive=true,                
    morekeywords={true,false,null},
    keywordstyle=\color{blue}\bfseries,
    stringstyle=\color{teal},
    commentstyle=\color{gray}\ttfamily,
    identifierstyle=\color{black},
    alsoletter={-},                
}

\newcommand{\bg}[2]{%
    \begingroup
    \setlength{\fboxsep}{3pt}%
    \colorbox{#1}{#2}%
    \endgroup
}

\newcommand{\rev}[1]{\textcolor{black}{#1}}

\newcommand{\bgp}[2]{%
    \begingroup
    \colorbox{#1}{%
        \begin{varwidth}{\dimexpr\linewidth-2\fboxsep\relax}%
            #2%
        \end{varwidth}%
    }%
    \endgroup
}

\definecolor{reason}{HTML}{D5E8D4}
\definecolor{speech}{HTML}{F8CECC}
\definecolor{trans}{HTML}{FFE6CC}
\definecolor{question}{HTML}{F5F5F5}

\title{\textsc{Stitch}: Simultaneous Thinking and Talking with Chunked Reasoning for Spoken Language Models}

\author{%
  Cheng-Han Chiang\textsuperscript{1,2}\thanks{Work done during an internship at Microsoft GenAI. \texttt{dcml0714@gmail.com}} \quad
  Xiaofei Wang\textsuperscript{2}\thanks{Correspondence: \texttt{xiaofei.wang@microsoft.com}} \quad
  Linjie Li\textsuperscript{2} \quad
  Chung-Ching Lin\textsuperscript{2} \quad
  Kevin Lin\textsuperscript{2} \quad\\
  \textbf{Shujie Liu}\textsuperscript{2} \quad
  \textbf{Zhendong Wang}\textsuperscript{2} \quad
  \textbf{Zhengyuan Yang}\textsuperscript{2} \quad
  \textbf{Hung-yi Lee}\textsuperscript{1} \quad
  \textbf{Lijuan Wang}\textsuperscript{2}\\[1ex]
  \textsuperscript{1}National Taiwan University \quad
  \textsuperscript{2}Microsoft
}

\iclrfinalcopy 
\begin{document}

\maketitle

\begin{abstract}
Spoken Language Models (SLMs) are designed to take speech inputs and produce spoken responses. 
However, current SLMs lack the ability to perform an internal, \textit{unspoken} thinking process before responding. 
In contrast, humans typically engage in complex mental reasoning internally, enabling them to communicate ideas clearly and concisely. 
Thus, integrating an unspoken thought process into SLMs is highly desirable.
While naively generating a complete chain-of-thought (CoT) reasoning before starting to talk can enable thinking for SLMs, this induces additional latency for the speech response, as the CoT reasoning can be arbitrarily long.
To solve this issue, we propose \textsc{\textbf{Stitch}}, a novel generation method that alternates between the generation of unspoken reasoning chunks and spoken response chunks.
Since the audio duration of a chunk of spoken response is much longer than the time to generate the tokens in a chunk of spoken response, we use the remaining free time to generate the unspoken reasoning tokens.
When a chunk of audio is played to the user, the model continues to generate the next unspoken reasoning chunk, achieving \textbf{simultaneous thinking and talking}.
Remarkably, \textsc{Stitch} matches the latency of baselines that cannot generate unspoken CoT by design while outperforming those baselines by 15\% on math reasoning datasets; \textsc{Stitch} also performs equally well on non-reasoning datasets as those baseline models.
Some animations and demonstrations are on the project page: \url{https://d223302.github.io/STITCH}.

\end{abstract}

\section{Introduction}

Spoken language models (SLMs) can take speech inputs and generate speech outputs~\citep{gpt4o,defossez2024moshi,arora2025landscape}.
SLMs generate speech by emitting some \textit{speech tokens}, discrete tokens that encode the information in speech~\citep{defossez2023high,zhang2024speechtokenizer}, which will be converted into an audio waveform by a speech decoder.
Most SLMs are trained from a backbone text-based LLM~\citep{defossez2024moshi,ding2025kimi}; we can obtain an SLM by teaching the backbone LLM to understand speech input and generate the speech tokens as the output~\citep{xiezhifei2024miniomni,llamaomni}.

Teaching the backbone LLM to directly generate speech tokens is challenging, as the LLM is originally trained to predict text tokens, so the speech tokens are completely new for the LLM backbone.
To make SLMs better predict the speech tokens, current mainstream methods first generate the text tokens before generating the speech tokens~\citep{chen2024slam,zeng2024glm,nguyen2025spirit}.
By design, those text tokens and the speech tokens are \textit{aligned}, in the sense that the text tokens correspond to the semantic content of the speech tokens.

While the aligned text-speech generation pipeline makes it easier to generate the speech response, these SLMs directly generate the output text-speech tokens without any inner thinking process, though explicitly modeling such reasoning steps has been shown helpful in text generation tasks~\citep{o1, guo2025deepseek}.
Humans can perform complex internal reasoning before we speak: we disambiguate questions, recall facts, manipulate variables, and then verbalize only the polished answer. 
Intuitively, this unspoken internal thinking helps improve the accuracy and coherence of our spoken content. 
However, verbalizing these internal thoughts can make communication confusing and unnecessarily lengthy.
Thus, it is desirable to enable SLMs a similar ability to think longer and silently before speaking out.

\begin{figure}[t!]
\includegraphics[clip, trim = 27px 15px 15px 10px, width = \textwidth]{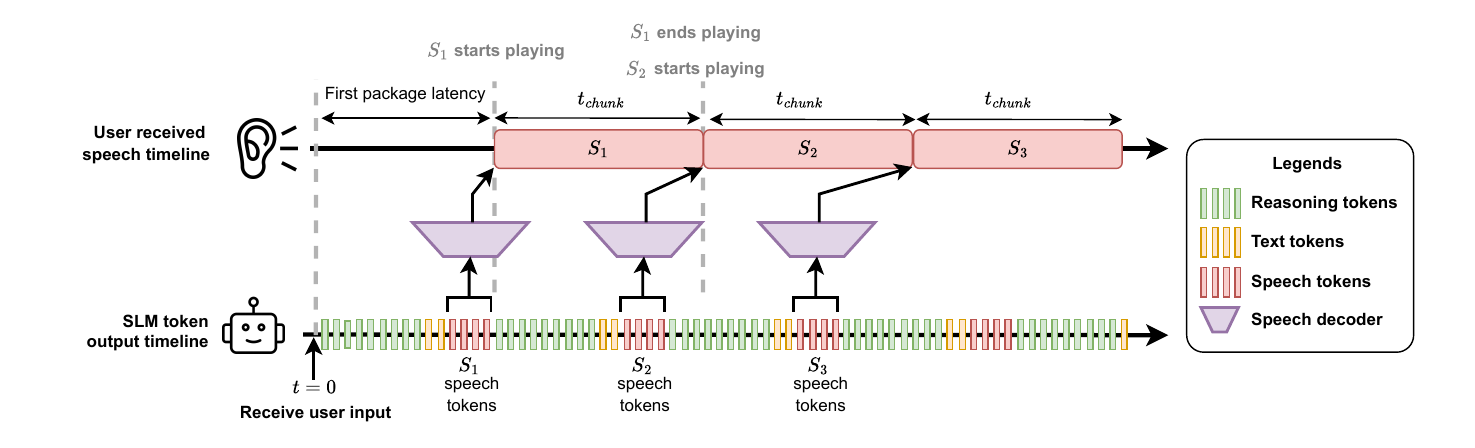}
\caption{
The timing diagram during generation for \textsc{Stitch-R}.
The model first generates the first $N_{\text{reason}}$ CoT reasoning tokens, $N_{\text{text}}$ text tokens, and $N_{\text{speech}}$ speech tokens.
Once the first $N_{\text{speech}}$ speech tokens are generated, the speech decoder synthesizes the output audio that lasts $t_{\text{chunk}}$ seconds.
When the speech waveform is synthesized and played to the user, the SLM uses this time to generate the next $N_{\text{reason}}$ reasoning tokens, $N_{\text{text}}$ text tokens, and $N_{\text{speech}}$ speech tokens and synthesize the speech output.
The duration $t_{\text{chunk}}$ is much longer than the time for generating the text tokens and speech tokens corresponding to $S_{i}$, and we use the remaining time to generate the reasoning tokens.} 
    \label{fig:fig1}
\end{figure}

A naive way to do so is to teach SLMs to generate an \textit{unspoken} chain-of-thought (CoT) reasoning~\citep{cot} in text, and then generate the spoken responses based on the CoT reasoning.
In this way, we can benefit from the CoT reasoning without making the spoken response overly long.
While this does improve the speech output quality, as validated by our experiments in Section~\ref{section: main results}, the speech responses need to wait until the full text CoT is generated, creating an uncontrollable latency for the speech output.

To reduce and remove this latency, we propose \textsc{Stitch}: \textbf{\underline{S}}imultaneous \underline{\textbf{T}}h\textbf{\underline{i}}nking and \textbf{\underline{T}}alking with \textbf{\underline{Ch}}unked Reasoning.
\textsc{Stitch} alternately generates unspoken reasoning chunks and spoken response chunks, allowing the spoken responses to condition on some thinking process.
Since the audio duration of a chunk of spoken response is much longer than the time to generate the tokens in a chunk of spoken response, we use the remaining free time to generate the unspoken reasoning tokens.
When a chunk of audio is played to the user, \textsc{Stitch} keeps generating the next unspoken reasoning chunk, achieving \textbf{simultaneous thinking and talking}.
\textsc{Stitch-S}, a variant of \textsc{Stitch}, has the same latency as the models that cannot generate unspoken reasoning, while significantly outperforming those baselines on math reasoning datasets.

We summarize our contributions as follows:\\
\noindent\hangindent=1.6em\hangafter=1\makebox[1.6em][l]{1.} We propose \textsc{Stitch}, an SLM generation pipeline that outputs unspoken reasoning chunks and spoken response chunks alternately.
To the best of our knowledge, we are the first to introduce the unspoken thinking process to SLMs.\\
2. \textsc{Stitch-S}, while improving the generation with unspoken reasoning, has a latency the same as an SLM that does not generate text reasoning by design.\\
3. On five math QA datasets, \textsc{Stitch} consistently outperforms the baseline models that cannot reason with text CoT.

\section{Related Work: Spoken Language Models}
\label{section: Related Work: Spoken Language Models}
Our paper improves SLMs by generating text reasoning before generating the speech output.
Current SLMs first generate some speech tokens, and the speech tokens will be synthesized into a waveform by the speech decoder.
In this section, we recap how existing SLMs generate speech tokens.
The mainstream SLMs nowadays adopt two types of speech generation methods: (1) \textbf{Interleaved} text-speech tokens~\citep{zeng2024glm,baichuanaudio} and the (2) \textbf{Thinker-Talker} architecture~\citep{qwen25omni,fang2025llama}.

In the \textbf{interleaved} decoding paradigm, the SLM backbone model generates a chunk of text tokens and a chunk of speech tokens alternately.
The text tokens serve as guidance for future speech tokens by transcribing what the speech token will say.
For example, GLM-4-Voice~\citep{zeng2024glm} interleaves between generating $N_{\text{text}}=13$ text tokens and $N_{\text{speech}}=26$ speech tokens.
After a chunk of speech tokens is generated, it is immediately synthesized into audio by the speech decoder and streamed to the user, enabling low latency and real-time interaction.
A figurative illustration of this output format is shown in Figure~\ref{fig:t2s2_generation_variants.pdf}(a).
When concatenating the chunk of text tokens, they should correspond to the transcription of the speech tokens.
The ratio of the text tokens and speech tokens is carefully selected such that the text tokens are always faster than the speech tokens to \textbf{ensure that the content of the speech tokens has already appeared in previous text tokens}.
Once all the text tokens are generated, the model will continue to generate the remaining speech tokens.

The \textbf{thinker-talker} SLM uses a \textit{thinker} model to take speech input and generate text tokens, and uses a \textit{talker} model to convert the text tokens and/or their continuous representations into speech tokens~\citep{qwen25omni,fang2025llama}.
The talker model functions like a Text-to-Speech (TTS) model that converts the textual output from the thinker into speech tokens, and the output by the thinker and talker will be highly aligned in their semantics, while there may be slight differences. 

The text-speech interleaved SLMs and thinker-talker SLMs differ in their architecture, but they share a similarity: they generate text and then the speech, and the text is highly aligned with the speech.
In this paper, we focus on enabling SLMs to generate an unspoken thinking process while speaking.

While reasoning, also sometimes referred to as thinking, is widely explored in text-only LLMs~\citep{o1,thinkingllm,guo2025deepseek,hong2025glm}, reasoning is only included in audio LLMs recently~\citep{wen2025sari,li2025reinforcement,audioreasoer}.
Audio LLMs are multimodal LLMs that can take audio and speech as input, but they cannot generate speech outputs~\citep{tang2023salmonn,chu2024qwen2,qwen25omni,lu2025desta2}; this marks a significant difference from SLMs studied in our paper.
Currently, no prior work introduces unspoken reasoning ability to SLMs.

\section{Method: Simultaneous Thinking and Talking with Chunked Reasoning}
\label{section: Method}

While current SLMs also include some textual inner monologue~\citep{defossez2024moshi} or thinking~\citep{qwen25omni} (as in thinker-talker) before generating the speech tokens, these text tokens simply transcribe what the model will say next.
In other words, they do not generate \textbf{\textit{unspoken}} reasoning before speaking.
In this paper, we aim to improve the SLMs by generating some unspoken text reasoning before generating the speech responses (including the text and speech tokens).

Our proposed concept can be applied to both the interleaved and thinker-talker SLMs, but \textbf{we adopt the interleaved SLMs} in this paper for the following \rev{three} reasons:
(1) The interleaved SLM only contains a single backbone model to generate the text and speech tokens, simplifying the model architecture. 
(2) \citet{wu2025towards} show that the thinker-talker SLM requires careful tuning to make it properly generate well-aligned text and speech output, making thinker-talker harder to fine-tune.
\rev{(3) Qwen-2.5-Omni, the most widely used thinker-talker SLM, does not publicly release the speech tokenizer, adding difficulties to fine-tune the talker model~\citep{qwenomniissue1,qwenomniissue2}.}

To unify the wording, we will use \textbf{\textit{reasoning tokens}} to refer to the \textbf{unspoken} reasoning (in text), \textbf{\textit{text tokens}} to refer to the original text tokens of the interleaved SLMs, and \textbf{\textit{speech tokens}} to refer to the speech tokens.
Text tokens and speech tokens are the output of interleaved SLMs from prior works~\citep{zeng2024glm}, and the reasoning tokens are introduced in this paper.
Both text tokens and reasoning tokens are texts, while they differ in their function: reasoning tokens are for the unspoken thinking process, and text tokens are the transcriptions of the spoken output.

\subsection{A Trivial Method: Think before Speaking in Speech (TBS)}
\label{subsection: T2S2: Thinking in Text before Speaking in Speech}
Before introducing \textsc{Stitch}, we start from a trivial baseline method that introduces the thinking process to SLMs: \textbf{T}hinking in text \textbf{B}efore \textbf{S}peaking in speech (TBS).
Introducing TBS can help us understand the motivation and design of \textsc{Stitch} later.
In TBS, given a user query in speech, the model first generates a complete trace of reasoning span.
A complete reasoning span is enclosed by two special tokens \texttt{[SOR]} and \texttt {[EOR]}, denoting the start and end of the reasoning span.
The \texttt{[EOR]} token lets the model know that the reasoning has finished and it needs to start generating the tokens for the spoken response, including the text tokens and the speech tokens.
The text tokens and the speech tokens are generated in an interleaved manner, following the original design of interleaved SLMs, with $N_{\text{text}}$ and $N_{\text{speech}}$ alternating with each other.

\begin{figure*}[t!]

\centering
\includegraphics[clip, trim = 27px 7px 15px 7px,width=0.98\linewidth]{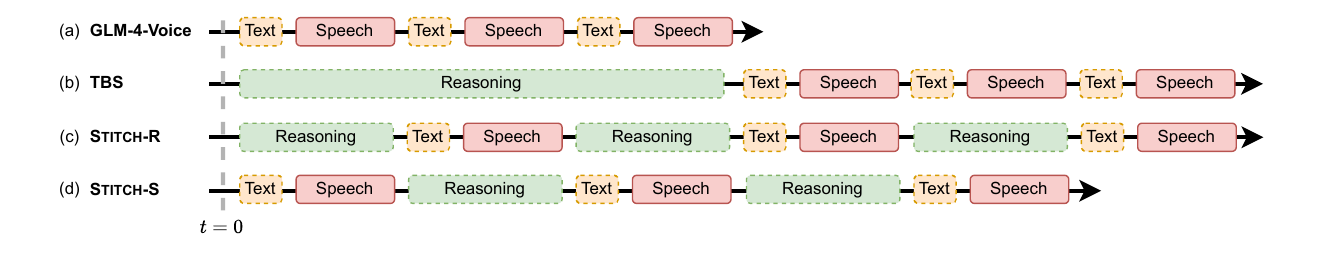}
\caption{Different generation method explored in this paper. 
The arrow represents the timeline for the SLM to generate the tokens; this timeline should \textbf{not be confused} with the timeline that the end user receives the audio, i.e., the upper timeline in Figure~\ref{fig:fig1}.
We plot tokens of the same type in a chunk using the same color.
(a) GLM-4-Voice: Interleaving between text and speech token chunks (Section~\ref{section: Related Work: Spoken Language Models}). 
This is the design of the original interleaved SLMs.
(b) TBS: Generating a complete reasoning span and then interleaving between text and speech token chunks (Section~\ref{subsection: T2S2: Thinking in Text before Speaking in Speech}).
(c) \textsc{Stitch}-R: Alternating between reasoning token chunks, text token chunks, and speech token chunks (Section~\ref{subsection: T2S2-Interleave: Interleaving Text Reasoning CoT and Speech Output}).
(d) \textsc{Stitch}-S: Alternating between text token chunks, speech token chunks, and reasoning token chunks (Section~\ref{subsection: T2S2-Reverse-Interleave: Speech Span First before Partial Reasoning Span}).}
\label{fig:t2s2_generation_variants.pdf}
\end{figure*}

To teach SLMs to operate in TBS, we construct the training data $\mathcal{D}_{\text{TBS}}$ where each training instance has the form $(\mathbf{x},\mathbf{z},\mathbf{y})$, $\mathbf{x}$ is the speech token sequence of the user input, $\mathbf{z}$ is the reasoning token sequence, and $\mathbf{y}=[\mathbf{t}_{1}\circ \mathbf{s}_{1} \circ \mathbf{t}_{2} \circ \mathbf{s}_{2}, \cdots]$ is the token sequence for the speech output that interleaves between $N_{\text{text}}$ text tokens ($\mathbf{t}_{j}$) and $N_{\text{speech}}$ tokens ($\mathbf{s}_{j}$)
(The last text token chunk may be less than $N_{\text{text}}$ while the last speech token span can have more than $N_{\text{speech}}$ tokens); $\circ$ denotes concatenating two token sequences.
We defer how we construct  $\mathcal{D}_{\text{TBS}}$ from existing datasets until Section~\ref{section: Experiment Setup: Training}.
A figurative illustration of the target output for TBS is in Figure~\ref{fig:t2s2_generation_variants.pdf}(b).

During training, given an input $\mathbf{x}$, we train the SLM to predict the reasoning tokens $\mathbf{z}$ and then the interleaved text-speech tokens $\mathbf{y}$ using the standard cross-entropy loss.

\subsection{\textsc{Stitch-R}: Simultaneous Thinking and Talking with Reasoning First}
\label{subsection: T2S2-Interleave: Interleaving Text Reasoning CoT and Speech Output}
While inserting text reasoning before the speech response in TBS improves the quality of the speech response, the speech response can only be generated after the full reasoning has been generated.
As later shown in Table~\ref{tab:merged-comparison}, the reasoning by TBS can be long, creating significant latency before generating the speech response.
To resolve the latency issue, we draw inspiration from the fact that humans can \textit{think while speaking} and propose \textsc{Stitch-R}.

\textsc{Stitch-R} realizes this "\textit{thinking when speaking}" by alternating fixed-length ($N_{\text{reason}}$) \textit{partial} reasoning spans, fixed-length ($N_{\text{text}}$) text token spans, and fixed-length ($N_{\text{speech}}$) speech token spans.
The partial reasoning spans are for inner thinking, while the text and speech token spans are for the spoken response.
Stitching the partial reasoning spans together will form a complete CoT reasoning.
A figurative illustration of the output of \textsc{Stitch-R} is shown in Figure~\ref{fig:t2s2_generation_variants.pdf}(c), and some samples generated by \textsc{Stitch-R} are shown in Table~\ref{tab:T2S2-I samples} in the Appendix. 
The "\textit{R}" in \textsc{Stitch-R} stands for \textit{"reasoning first"} since it generates a partial reasoning chunk before speaking; this is used to distinguish the "\textit{speaking first}" \textsc{Stitch-S} that will be introduced in Section~\ref{subsection: T2S2-Reverse-Interleave: Speech Span First before Partial Reasoning Span}.

We explain how simultaneous thinking and speaking is achieved by \textsc{Stitch-R}, which can be most easily understood with Figure~\ref{fig:fig1}.
After the SLM receives the speech input, it first generates $N_{\text{reason}}$ reasoning tokens, $N_{\text{text}}$ text tokens, and $N_{\text{speech}}$ speech tokens for the first audio output chunk $S_1$.
Those speech tokens will be passed through the speech decoder to synthesize the audio with duration $t_{\text{chunk}}$ seconds and played to the user.
When synthesizing the speech waveform and playing the audio for $S_1$, the SLM continues to generate the next $N_{\text{reason}}$ reasoning tokens; \textbf{this is where simultaneous thinking and speaking happens.}
After generating the second reasoning chunk, the model continues to generate $N_{\text{text}}$ text tokens, and $N_{\text{speech}}$ speech tokens for the next audio chunk $S_2$.
As long as the time to generate $N_{\text{reason}}+N_{\text{text}}+N_{\text{speech}}$ tokens, denoted as $t_{token}$, is less than $t_{\text{chunk}}$, the next audio chunk $S_2$ can be played seamlessly after the previous audio chunk $S_1$.

Let's understand with some concrete numbers.
On an A100-80G GPU, our model generate around 80 tokens per second using vLLM~\citep{kwon2023efficient}, and the duration of the audio for $N_{\text{speech}}=26$ speech tokens is around $t_{\text{chunk}}=2$ seconds, so we can generate around 160 tokens in $t_{\text{chunk}}=2$ seconds. 
The number of text tokens and speech tokens sums to $N_{\text{text}}+N_{\text{speech}}=13+26=39$, so we have time to generate at most $80\times2-39=121$ reasoning tokens.\footnote{Since the token generation and audio synthesis can be run in parallel, we do not need to consider $t_{\text{syn}}$, the time to synthesize the audio, as long as $t_{\text{syn}}<t_{\text{chunk}}$.}

Each partial reasoning chunk is enclosed in two special tokens: \texttt{[SOPR]} (start of partial reasoning) and \texttt{[EOPR]} (end of partial reasoning).
If the reasoning chunk is fully generated, an additional \texttt{[EOR]} is used to denote the end of the full reasoning CoT.

To construct the training data for \textsc{Stitch-R}, we simply split the full reasoning CoT $\mathbf{z}$ in $\mathcal{D}_{\text{TBS}}$ into chunks with $N_{\text{reason}}$ tokens $\{\mathbf{z}_{1}, \mathbf{z}_{2},\cdots\}$, where each $\mathbf{z}_{i}$ except the last chunk has $N_{\text{reason}}$ tokens.
Next, we interleave those chunks with the interleaved text-speech token sequence $\mathbf{y}=[\mathbf{t}_{1}\circ \mathbf{s}_{1} \circ \mathbf{t}_{2} \circ \mathbf{s}_{2}, \cdots]$ to create the interleaved data of the form $[\mathbf{z}_{1}\circ \mathbf{t}_{1}\circ \mathbf{s}_{1} \circ \mathbf{z}_{2} \circ \mathbf{t}_{2} \circ \mathbf{s}_{2} \circ \cdots]$, as shown in Figure~\ref{fig:t2s2_generation_variants.pdf}(c).
If the number of the reasoning span is more than the number of text spans, this indicates that the reasoning token spans \textit{think slower} than the text token spans, so we remove the sample from the training data.\footnote{The number of text spans is always less than or equal to the number of speech spans by design. This can be done by properly selecting the ratio between $N_{\text{text}}$ and $N_{\text{speech}}$.}
The model is fine-tuned to auto-regressively predict the interleaved reasoning-text-speech token spans using standard language modeling cross-entropy loss.

\subsection{\textsc{Stitch-S}: Simultaneous Thinking and Talking with Speaking First}
\label{subsection: T2S2-Reverse-Interleave: Speech Span First before Partial Reasoning Span}

\textsc{Stitch-R} can have a latency shorter than that of TBS, since \textsc{Stitch-R} only needs to wait at most $N_{\text{reason}}$ tokens before generating the text and speech tokens.
However, compared with the original text-speech interleaved SLMs that directly generate the text and speech tokens for the spoken response, \textsc{Stitch-R} still needs to wait for $N_{\text{reason}}$ reasoning tokens.

To fully remove the latency of waiting for the first partial reasoning span, we propose an alternative generative pipeline that directly starts to generate the text and speech token chunks and then generates the first reasoning chunk; the model continues to interleave this generation pattern.
We call this \textsc{Stitch-S} since it generates a speech response first, and an illustrative figure is shown in Figure~\ref{fig:t2s2_generation_variants.pdf}(d).

To construct the training data for \textsc{Stitch-S}, we similarly split the full reasoning $\textbf{z}$ in $\mathcal{D}_{\text{TBS}}$ into chunks with $N_{\text{reason}}$ tokens, and insert a text token chunk of length $N_{\text{text}}$ and a speech token chunk of length $N_{\text{speech}}$ \textbf{before} each reasoning span.
Following previous notation, the target output for \textsc{Stitch-S} looks like $[\mathbf{t}_{1}\circ \mathbf{s}_{1} \circ \mathbf{z}_{1} \circ \mathbf{t}_{2} \circ \mathbf{s}_{2} \circ \mathbf{z}_{2} \circ \mathbf{t}_{3} \circ \mathbf{s}_{3}, \cdots]$.
If a training instance's number of reasoning chunks is more than the number of text chunks, the above data construction pipeline will result in an instance that has a reasoning span after the text chunks are finished, which does not make sense since the thinking process hasn't ended while the content to be spoken is finished.
We remove this kind of sample from the training dataset.

\section{Experiment Setup}

\subsection{Training}
\label{section: Experiment Setup: Training}
We fine-tune a separate model for TBS, \textsc{Stitch-R}, and \textsc{Stitch-S} from GLM-4-Voice-9B~\citep{zeng2024glm}.
During fine-tuning, we fully fine-tune the backbone transformer model~\citep{vaswani2017attention} of GLM-4-Voice and freeze the speech encoder~\citep{zeng2025scaling} and speech decoder~\citep{du2024cosyvoice,kong2020hifi}.
The detailed hyperparameters for fine-tuning are listed in Appendix~\ref{Appendix: section: Training Details}. 

We include three types of tasks in the training data: general dialogue dataset (VoiceAssistant400K~\citep{xiezhifei2024miniomni}), mathematical reasoning (Tulu-3-Persona Math, Tulu-3-Persona-Math-Grade, and Tulu-3-Persona-Algebra~\citep{lambert2024tulu}), and knowledge-intensive question answering (Natural Question~\citep{kwiatkowski-etal-2019-natural} and TriviaQA~\citep{joshi-etal-2017-triviaqa}).

We introduce how to construct $\mathcal{D}_{\text{TBS}}$, the training data for TBS, while the training data for \textsc{Stitch-R} and \textsc{Stitch-S} can be constructed from  $\mathcal{D}_{\text{TBS}}$ following the procedures in Section~\ref{subsection: T2S2-Interleave: Interleaving Text Reasoning CoT and Speech Output} and \ref{subsection: T2S2-Reverse-Interleave: Speech Span First before Partial Reasoning Span}.

To train TBS, each sample in the dataset needs to have the spoken query $\mathbf{x}$, text reasoning $\mathbf{z}$, and the spoken responses $\mathbf{y}$.
For VoiceAssistant400K, which is already a speech-to-speech dialogue dataset, we generate the text reasoning using GPT-4o.
For mathematical reasoning datasets, which only contain a question and its long answer in text, we synthesize the question audio $\mathbf{x}$ using GPT-4o-mini-TTS~\citep{4ominitts}, use the ground truth long answer as the text CoT $\mathbf{z}$, and rewrite the long answer and synthesize its audio to form the spoken response $\mathbf{y}$.
For knowledge-intensive QA datasets, which are also in text form, we synthesize the question audio $\mathbf{x}$ using GPT-4o-mini-TTS, generate the text CoT $\mathbf{z}$ with GPT-4o, and rewrite the answer and synthesize its audio to form the spoken response $\mathbf{y}$.
All the prompts used for constructing the datasets are in Appendix~\ref{appendix: subsection: Training Dataset}.
The full training dataset contains around 400K instances.

\subsection{Evaluation}
\label{subsection: Training Data}

We include reasoning and non-reasoning datasets for evaluation.
We use the former to evaluate whether reasoning before responding in speech can improve the spoken answer quality for reasoning tasks.
We include five math reasoning datasets: AddSub~\citep{hosseini2014learning}, SinglEq~\citep{koncel2015parsing}, MultiArith~\citep{roy-roth-2015-solving}, SVAMP~\citep{patel2021nlp}, and GSM8K~\citep{cobbe2021training}, which are commonly used to evaluate the math reasoning ability of LLMs~\citep{cot,kojima2022large,wang-etal-2023-plan}.
The questions and answers in these datasets are not too difficult and can be described clearly in speech.
Since these datasets are originally in textual form, we synthesize the question audio using GPT-4o-mini-TTS.

For non-reasoning datasets, we include three knowledge-intensive QA datasets (Llama Questions~\citep{nachmani2024spoken}, TriviaQA~\citep{joshi-etal-2017-triviaqa}, and WebQuestions~\citep{berant-etal-2013-semantic}) and AlpacaEval~\citep{alpaca_eval}, a dialogue dataset, from OpenAudioBench~\citep{baichuanaudio}.
We use these datasets to understand whether generating unspoken reasoning before responding in speech has any impact on non-reasoning datasets.
The dataset statistics are in Appendix Table~\ref{tab:test datasets}.

For all reasoning datasets and knowledge-intensive QA datasets, we evaluate the SLM's performance using the accuracy.
To calculate the accuracy, we use GPT-4o to judge whether the SLM's prediction matches the ground truth answer.
For AlpacaEval, we follow the setting in OpenAudioBench to use LLM-as-a-judge~\citep{chiang-lee-2023-large,liu-etal-2023-g,zheng2023judging} to rate the response from 1 to 10, and report the GPT-4o-score scaled to 10 and 100, where higher is better.

When evaluating the responses from SLMs, we use the text tokens for evaluation.
Alternatively, we can synthesize the audio from the speech tokens, transcribe the audio, and evaluate those transcriptions.
However, since the SLM's text tokens are designed to align with the speech output, we only report the performance of the text tokens here.
The general trend observed in the following experiment results holds when we evaluate the transcription, while the latter has a slight performance drop of around 3\% to 6\%, as shown in Table~\ref{tab:reasoning asr} in the Appendix.

\subsection{Compared Models}
\label{subsection: Compared Models}

We compare six models, which are based on the same base model: GLM-4-Voice, an interleaved SLM.
GLM-4-Voice has an $N_{\text{text}}=13$ and $N_{\text{speech}}=26$.
All the models we use follow the $N_{\text{text}}$ and $N_{\text{speech}}$ in GLM-4-Voice.

(1) \textbf{GLM-4-Voice}: The model takes speech input and generates interleaved text-speech tokens.
This model cannot generate unspoken reasoning by design. \quad
(2) \textbf{No reasoning}: We fine-tune GLM-4-Voice on a dataset without reasoning, $\mathcal{D}_{\text{no-reason}}$.
Each instance in $\mathcal{D}_{\text{no-reason}}$ has the form $(\mathbf{x}, \mathbf{y})$, which is constructed from $\mathcal{D}_{\text{TBS}}$ by removing the text reasoning $\mathbf{z}$. \quad
The model is trained to predict interleaved text-speech tokens $\mathbf{y}$ based on the input $\mathbf{x}$. \quad
(3) \textbf{TBS}: see Section \ref{subsection: T2S2: Thinking in Text before Speaking in Speech}.\quad
(4) \textbf{Mix reasoning}: We mix half of $\mathcal{D}_{\text{no-reason}}$ and half of $\mathcal{D}_{\text{TBS}}$ to fine-tune GLM-4-Voice.
We use a system prompt to control whether the model should generate reasoning before generating the interleaved text-speech tokens.
This allows us to control the model's behavior during inference. \quad
By comparing the performance of this model when inference with and without reasoning, we can understand if unspoken reasoning really helps SLM's speech output.\quad
(5) \textbf{\textsc{Stitch-R}}: see Section \ref{subsection: T2S2-Interleave: Interleaving Text Reasoning CoT and Speech Output}.\quad 
(6) \textbf{\textsc{Stitch-S}}: see Section \ref{subsection: T2S2-Reverse-Interleave: Speech Span First before Partial Reasoning Span}.

During training, we set $N_{\text{reason}}=100$ for \textsc{Stitch-R} and \textsc{Stitch-S}.
In Section~\ref{subsection: T2S2-Interleave: Interleaving Text Reasoning CoT and Speech Output}, we have derived that the length of the reasoning token can be at most 121 tokens when running on an A100-80G GPU, so we select $N_{\text{reason}}=100$.
We do not compare with other SLMs as they use different training data and have distinct model architectures, making the comparison with our models less meaningful.


\begin{table}[t!]
    \centering

    \begin{subtable}[t]{\textwidth}
        \centering
        \caption{The accuracy on math QA.}
        \label{tab:math-reasoning}
        \setlength{\tabcolsep}{4pt}
        \renewcommand{\arraystretch}{1.2}
        \footnotesize
        \begin{adjustbox}{max width=\textwidth}
            \begin{tabular}{ccccc|ccccc|c}
                \toprule
                
                 \multirow{2}{*}{Id}&\multirow{2}{*}{Config}   
                   &\multirow{2}{*}{Latency}& \multicolumn{2}{c}{\textit{Use Reasoning}}  
                  & \multicolumn{5}{|c|}{\textit{Datasets}} 
                  & \multirow{2}{*}{\textit{Avg.}} \\
                   & 
                   &                   & Train  
                  & Infer.  
                  & \textit{AddSub} 
                  & \textit{MultiArith} 
                  & \textit{SinglEq} 
                  & \textit{SVAMP} 
                  & \textit{GSM8K} 
                  &                     \\
                \midrule
                 &\multicolumn{9}{c}{\textit{\textbf{Baselines}}} &  \\
                \midrule
                 1&GLM-4-Voice             
                   &$N_{\text{text}}+N_{\text{speech}}$& –       
                  & \ding{56} 
                  & 59.42           
                  & 62.00               
                  & 71.00            
                  & 44.00          
                  & 29.00         
                  & 53.08               \\
                 2&No reasoning            
                   &$N_{\text{text}}+N_{\text{speech}}$& \ding{56}
                  & \ding{56}
                  & 66.06           
                  & 70.69               
                  & 77.98            
                  & 64.43          
                  & 35.73         
                  & 62.98               \\
                  3&Mix reasoning
                   &$N_{\text{text}}+N_{\text{speech}}$& \ding{52}+\ding{56}
                  & \ding{56}
                  & 69.72           
                  & 79.31               
                  & 87.16            
                  & 63.55          
                  & 37.33         
                  & 67.41               \\
                  \midrule
                   &\multicolumn{9}{c}{\textit{\textbf{Full Reasoning before Speaking}}} &  \\
                  \midrule
                  4&TBS                  
                   &$N_{\text{full}}+N_{\text{text}}+N_{\text{speech}}$& \ding{52}
                  & \ding{52}
                  & 79.82           
                  & 85.63               
                  & 89.91            
                  & 75.29          
                  & 64.94         
                  & 79.12               \\
                
                  5&Mix reasoning
                   &$N_{\text{full}}+N_{\text{text}}+N_{\text{speech}}$& \ding{52}+\ding{56}         
                  & \ding{52}
                  & 76.15           
                  & 89.08               
                  & 88.99            
                  & 70.90          
                  & 62.23         
                  & 77.47               \\
                \midrule
                 &\multicolumn{9}{c}{\textit{\textbf{Proposed Methods: Simultaneous Thinking and Speaking}}} &  \\
                \midrule

                 6&\textsc{Stitch-R}                  
                   &$N_{\text{reason}}+N_{\text{text}}+N_{\text{speech}}$& \ding{52}
                  & \ding{52}
                  & 78.90           
                  & 88.51               
                  & 93.58            
                  & 73.83          
                  & 58.70         
                  & 78.70               \\
                 7&\textsc{Stitch-S}                
                   &$N_{\text{text}}+N_{\text{speech}}$& \ding{52}
                  & \ding{52}
                  & 81.65           
                  & 87.93               
                  & 91.74            
                  & 72.15          
                  & 56.72         
                  & 78.04               \\
                \bottomrule
            \end{tabular}
        \end{adjustbox}
    \end{subtable}

    \begin{subtable}[t]{\textwidth}
        \centering
        \caption{The performance on non-reasoning tasks.
        For factual QA, including Llama Question, TriviaQA, and WebQuestion, we report the accuracy.
        For AlpacaEval, we report the GPT-4o score.}
        \label{tab:non-reasoning}
        \footnotesize
        \begin{adjustbox}{max width=\textwidth}
            \begin{tabular}{cccc|cccc|c}
                \toprule
                 \multirow{2}{*}{Id.}&\multirow{2}{*}{Config}   
                  & \multicolumn{2}{c}{\textit{Use Reasoning}}  
                  & \multicolumn{4}{|c|}{\textit{Datasets}} 
                  & \multirow{2}{*}{\textit{Avg.}} \\
                                &
                  & Train  
                  & Infer.  
                  & \textit{Llama Question} 
                  & \textit{TriviaQA} 
                  & \textit{WebQuestion} 
                  & \textit{AlpacaEval} 
                  &                     \\
                \midrule
                 &\multicolumn{8}{c}{\textit{\textbf{Inference without Reasoning}}}  \\
                \midrule
                 1&GLM-4-Voice             
                  & –       
                  & \ding{56} 
                  & 74.25            
                  & 47.10             
                  & 50.95          
                  & 48.59               
                  & 55.22               \\
                 2&No reasoning            
                  & \ding{56}
                  & \ding{56}
                  & 73.33            
                  & 46.00             
                  & 47.80          
                  & 53.62               
                  & 55.19               \\
                    3&Mix reasoning
                  & \ding{52}+\ding{56}
                  & \ding{56}
                  & 73.33            
                  & 47.80             
                  & 48.00          
                  & 54.77               
                  & 55.98               \\
                  \midrule
                   &\multicolumn{8}{c}{\textit{\textbf{Full Reasoning before Speaking}}}  \\
                  \midrule
                   4&TBS                   
                  & \ding{52}
                  & \ding{52}
                  & 74.33            
                  & 51.50             
                  & 52.15          
                  & 56.33               
                  & 58.58               \\
               
                   5&Mix reasoning
                  & \ding{52}+\ding{56}
                  & \ding{52}
                  & 69.67            
                  & 48.80             
                  & 50.75          
                  & 56.03               
                  & 56.31               \\
                \midrule
                 &\multicolumn{8}{c}{\textit{\textbf{Proposed Methods: Simultaneous Thinking and Speaking}}}  \\
                \midrule
                
                 6&\textsc{Stitch-R}               
                  & \ding{52}
                  & \ding{52}
                  & 70.00            
                  & 49.60             
                  & 50.25          
                  & 54.02               
                  & 55.97               \\
                 7&\textsc{Stitch-S}                
                  & \ding{52}
                  & \ding{52}
                  & 73.33            
                  & 50.00             
                  & 50.15          
                  & 54.62               
                  & 57.03               \\
                \bottomrule
            \end{tabular}
        \end{adjustbox}
    \end{subtable}

    \caption{Performance on reasoning dataset (\ref{tab:math-reasoning}) and non-reasoning datasets (\ref{tab:non-reasoning}).
    The \textit{latency} column reports how many tokens the SLM should generate until the first chunk of speech can be synthesized, where $N_{\text{full}}$ denotes the uncontrollable length of the full reasoning.
    The \textit{"use reasoning"} columns denote whether text reasoning $\mathbf{z}$ is in the training data or is generated during inference.}
    \label{tab:reasoning-vs-non}
\end{table}

\section{Main Results}
\label{section: main results}

We present the main results in Table~\ref{tab:reasoning-vs-non}.
We summarize several key observations as follows:

\textbf{TBS significantly outperforms the baselines without reasoning on math QA.}\quad
In row 4, TBS achieves an average accuracy of 79.12\%, outperforming GLM-4-Voice (row 1) by 26.04\% and the "\textit{no reasoning}" model (row 2) by 16.14\%.
Most significantly, the accuracy of TBS almost doubles compared to the two baselines without reasoning on GSM8K.
\rev{Using McNemar's test, we found that the TBS outperforms GLM-4-Voice and the no-reasoning baselines in a statistically significant way, with $p$-value less than $0.05$.}
This shows that reasoning before speaking can help the SLM respond better to questions that require reasoning.

\begin{table}[t!]
    \centering
    \setlength{\tabcolsep}{4pt}
    \renewcommand{\arraystretch}{1.2}
    \footnotesize
    \begin{adjustbox}{max width=\textwidth}
        \begin{tabular}{c|cccccc|cc|cc}
            \toprule
            & 
            \multicolumn{6}{c|}{\textbf{\underline{Token Count}}} 
            & 
            \multicolumn{4}{c}{\textbf{\underline{Quality Assesment}}}\\[4pt]
            Datasets  & \multicolumn{2}{c}{MultiArith}
              & \multicolumn{2}{c}{GSM8K}
              & \multicolumn{2}{c|}{TriviaQA}
              & MultiArith & GSM8K
              & MultiArith & GSM8K
               \\
            Metrics & Reason & Trans
            & Reason & Trans
            & Reason & Trans
            & \multicolumn{2}{c|}{UTMOSv2}
            & \multicolumn{2}{c}{GPT-4o-score}
             \\
            \midrule
            GLM-4-Voice
              & --     &  66.17
              & --     &  92.38
              & --     &  51.67
              &  3.10   &  3.05
              &  4.60     &  4.28 \\
            No reasoning
              & --     &  45.67
              & --     &  60.22
              & --     &   28.64
              &  3.15   &  3.13
              &  4.91     &  4.49 \\
            TBS
              & 178.95 &  45.09
              & 360.04 &  70.49
              &  68.03 &  27.55
              &  3.14   &  3.16
              &  4.90     &  4.78 \\
            \textsc{Stitch-R}
              & 157.32 &  49.77
              & 322.40 &  74.36
              &  67.56 &  27.65
              &  3.10   &  3.12
              &  4.94     &  4.74 \\
            \textsc{Stitch-S}
              & 203.28 &  80.87
              & 334.71 &  61.85
              &  69.45 &  28.21
              &  3.17   &  3.11
              &  4.88     &  4.66 \\
            \bottomrule
        \end{tabular}
    \end{adjustbox}
    \caption{Left: The average number of total reasoning tokens (summed over partial reasoning spans for \textsc{Stitch-R} and \textsc{Stitch-S}) and the text tokens.
    Right: The quality scores of the speech responses.}
    \label{tab:merged-comparison}
\end{table}

\textbf{Reasoning before speaking is better than no reasoning before speaking, even if the training data is the same.}\quad
The strong result of TBS, compared with the "\textit{no reasoning}" baseline, may raise a question on whether TBS is winning because the training data includes additional reasoning $\mathbf{z}$, which are not included in the training data for the "\textit{no reasoning}" baseline.
This question can be answered by comparing the results of the "\textit{mix reasoning}" model with and without text reasoning before generating the speech response (row 3 vs. row 5).
Since the two different inference modes use the same \textit{mix reasoning} model, the performance difference can be attributed to the reasoning before the spoken response during inference.
On average, reasoning before speaking (row 5) obtains an accuracy of 77.47\%, outperforming not using reasoning by more than 10\% (row 3).
For the most difficult GSM8K, the improvement of using reasoning is 24.9\%.
The results again confirm the importance of generating reasoning before the speech response for reasoning tasks.

\textbf{\textsc{Stitch-R} (row 6) shows almost no performance drop on average.}\quad
\textsc{Stitch-R} alternates between partial text reasoning and text-speech tokens.
While the reasoning is split by text-speech tokens, the average performance only drops by 0.42\% compared to TBS.
In Table~\ref{tab:merged-comparison}, we report the average number of reasoning tokens on three testing datasets.
Both TBS and \textsc{Stitch-R} generate 150 to 300 reasoning tokens for math reasoning datasets on average, but \textsc{Stitch-R} only needs to wait for the first $N_{token}=100$ reasoning tokens before starting to generate the first chunk of the speech output, yielding a smaller latency for \textsc{Stitch-R}.
Importantly, this speedup comes with almost no loss in accuracy.
Still, on a more difficult dataset (GSM8K), the performance drop is larger, but it still significantly outperforms the non-reasoning baselines (row 1 and row 2).
\rev{Again, using McNemar's test, we find that the STITCH-R outperforms the no-reasoning baselines in a statistically significant way.}
We discuss two samples from \textsc{STITCH-R} in Section~\ref{subsection: Qualitative Results}.

\textbf{\textsc{Stitch-S} shows almost no performance drop on average without adding any inference latency.}\quad
Remarkably, \textsc{Stitch-S} shows an average accuracy of 78.04\% (row 7), slightly lower than TBS by 1.08\% (row 4), but is still much better than the "\textit{no reasoning}" baseline by 15.06\% (row 2).
We see a larger performance drop in GSM8K around 8\% compared with TBS, but again, still significantly outperforms \textit{no reasoning} baseline by 19.99\%\rev{, again in a statistically significant way based on McNemar's test}.
Recall that the first packet latency of \textsc{Stitch-S} is exactly the same as the baselines that do not generate text reasoning; both of them only need to wait for $N_{\text{text}}+N_{\text{speech}}$ tokens before synthesizing the speech waveform.
This exciting result indicates that we can improve the answer quality \textbf{without sacrificing any inference latency}.
By scrutinizing the responses from \textsc{Stitch-S}, we find that the first text token chunk is often a rephrase of the question, which can be generated without any reasoning.
This explains why \textsc{Stitch-S} does not show a severe accuracy drop compared with \textsc{Stitch-R}.

\textbf{On non-reasoning datasets, reasoning models perform comparably well or slightly outperform non-reasoning baselines.}\quad
Table~\ref{tab:non-reasoning} shows the performance on non-reasoning datasets.
We find that TBS (row 4), \textsc{Stitch-R} (row 6), and \textsc{Stitch-S} (row 7) all have performance comparable to or better than the GLM-4-Voice (row 1) and the \textit{no reasoning} baseline (row 2).
This shows that fine-tuning SLMs to think in text before responding does not harm the performance on non-reasoning datasets.

\textbf{Generating text reasoning before generating the speech responses does not harm the perceptual quality and the fluency of the speech response.}\quad
In the rightmost block in Table~\ref{tab:merged-comparison}, we report some quality assessment of the responses generated by the SLMs.
We report (1) the perceptual quality of the speech response evaluated by UTMOSv2~\citep{baba2024t05}, an automatic MOS evaluation model, and (2) the fluency score of the text tokens judged by GPT-4o~\citep{chiang-lee-2023-large}.
The fluency score focuses on whether the text tokens form a fluent narrative.
Both metrics range from 1 to 5, with 5 being the best.
We observe that all fine-tuned models have a similar UTMOSv2 score over 3, similar to the original GLM-4-Voice, indicating that our fine-tuning does not harm the perceptual quality of the speech output. 
We also observe that the fluency scores for \textsc{Stitch-R} and \textsc{Stitch-S} are high, indicating that alternating between the text reasoning token spans and the text-speech token spans does not harm the fluency of the text tokens. 

\rev{
\textbf{Human evaluation shows that STITCH-S is more responsive than STITCH-R, followed by TBS.}\quad
We conduct a human evaluation on the responsiveness of different models.
We present the human evaluator with a question from the user and two audio responses from different models, asking them to compare the responsiveness of the two models on a seven-point scale, where +3 indicates that model A is more responsive than model B and -3 indicates that model B is more responsive than model A.
To simulate the first packet latency that the user may experience, each audio includes a silence in the beginning, accounting for the prefilling and decoding time of the SLM and the speech synthesis latency.
We compare the following pairs: TBS / STITCH-R, TBS / STITCH-S, STITCH-R / STITCH-S, and STITCH / no-reasoning.
For each pair, we use the same 100 questions from GSM8K, and we recruit 3 human evaluators to judge the responsiveness of a pair of models.
The results are presented in Table~\ref{tab: human evaluation} in the Appendix, and we summarize the results as follows:
1. STICH-R and STITCH-S are both more responsive than TBS.
2. STITCH-S is more responsive than STITCH-R.
3. STITCH-S and no-reasoning are equally responsive.}

\rev{
\textbf{The reasoning segments finish much earlier than the text segments.}\quad
Based on the statistics in Table~\ref{tab:merged-comparison}, we can see that the average number of reasoning chunks are less than the number of text chunks.
For example, on GSM8K, STITCH-R has on average $322.40/100 =3.22$ reasoning chunks, while the average number of text chunks is $74.36/13=5.72$.
During inference time, we do not find the interleaved text-speech chunks to end before the reasoning chunks finish, i.e., before the \texttt{EOR} is emitted.
This is mainly because when we construct the training data for STITCH-R and STITCH, we remove the samples whose reasoning chunks are more than the text chunks, i.e., those samples that have reasoning slower than the response.
As a result, after training, the model will learn that it must finish the reasoning before concluding its spoken response.}

\section{\rev{Qualitative Results}}
\label{subsection: Qualitative Results}

In this section, we provide two outputs from our \textsc{Stitch-R} in Table~\ref{tab:T2S2-I samples} and a sample for \textsc{Stitch-S} in Table~\ref{tab:T2S2-IR samples} in the Appendix.
The top sample in Table~\ref{tab:T2S2-I samples} shows a typical pattern of the output from \textsc{Stitch-R}, where all the calculation results from the text token chunks are already calculated in previous reasoning chunks.
When generating the text tokens, the model simply needs to rewrite the lengthy reasoning into more concise sentences that are easier for spoken communication.

The bottom sample in Table~\ref{tab:T2S2-I samples} shows a somewhat rare pattern, where the second text token chunk includes a new calculation result (the number 120) that has not been explicitly calculated in previous reasoning chunks.
This shows that, besides thinking during the reasoning chunks, the model sometimes also incorporates the text tokens for thinking.
However, this pattern is rare: We use GPT-4o to parse the 1310 responses to GSM8K and find that only 5\% of the samples illustrate this pattern. 
The same pattern is also rare on \textsc{Stitch-S}.

Last, we find that the first text token chunk tends to rephrase the information of the question without adding new calculation results, which can be seen in the two samples in Table~\ref{tab:T2S2-I samples}.
This shows that the first text token chunk can be generated without relying on additional reasoning.
This can explain why \textsc{Stitch-S} can have a performance similar to TBS and \textsc{Stitch-R}: since the first text token chunk can be generated without thinking, we can directly generate those texts without conditioning on any reasoning.

\section{Adjusting the Length of the Reasoning Chunk During Inference}
\label{subsection: Adjust Length}
During training, we fix $N_{token}$, the number of tokens in each partial reasoning chunk, to 100 tokens.
In this section, we ask whether we can adjust $N'_{token}$ at test time without retraining the model.
One might want to adjust $N'_{token}$ at test time if the hardware is slower and the time to generate $N_{token}+N_{\text{text}}+N_{\text{speech}}$ tokens may exceed the duration $t_{\text{chunk}}$ of the audio for $N_{\text{speech}}$ tokens.
In this case, the model deployer may want to reduce $N'_{token}$ during inference time to ensure that generating the $N'_{token}+N_{\text{text}}+N_{\text{speech}}$ does not exceed the duration $t_{\text{chunk}}$.
Thus, it would be interesting to see if we can train a single model with a fixed $N_{token}$ and vary $N'_{token}$ during inference.

Recall that each reasoning chunk ends with \texttt{[EOPR]}.
To adjust the length of the reasoning chunk to $N'_{token}\le N_{token}$ during the inference time, we can append \texttt{[EOPR]} after $N'_{token}$ reasoning tokens are generated.
Once the model sees \texttt{[EOPR]}, it will cease generating the reasoning tokens and switch to generating text-speech interleaved tokens.
We intervene in the model's generation by the above process whenever the model generates a reasoning chunk until the \texttt{[EOR]} token is generated.

We vary the number of $N'_{token}$ from 60 to 100 with a 10 increment on the math reasoning datasets for \textsc{Stitch-R} and \textsc{Stitch-S}.
The results are shown in Appendix Figure~\ref{fig:Images/T2S2-I.pdf} and \ref{fig:Images/T2S2-RI.pdf}; we summarize the main findings below.
We see that the accuracy on all tasks mostly rises when increasing $N'_{token}$ from 60 to 100.
While the performance of $N'_{token}<100$ is mostly worse than the performance of $N_{token}=100$, using $N'_{token}<100$ is still better than the "\textit{no reasoning}" baseline when $N'_{token}\geq70$.
We also observe that when $N'_{token}\geq80$ tokens, the accuracy recovers to 90\% of the accuracy of $N_{token}=100$.
The results indicate that the model deployer can control the number of $N'_{token}$ in a reasonable range during inference without retraining the model with some accuracy trade-off.

\section{Using Reasoning from Other Models}
\label{subsection: Using text reasoning from Other Models}

Previously, we showed that \textsc{Stitch-R} and \textsc{Stitch-S} can generate reasoning before responding in speech.
Here, we discuss whether we can improve the speech response by using the reasoning from \textit{other} models.
The motivation is that sometimes the reasoning generated by the SLM itself is not good enough or takes too long to generate, and we may use other stronger or smaller models to generate the reasoning.
We call the model used to generate the reasoning the \textit{"reasoning augmentation model"}.

We discuss how to improve \textsc{Stitch-R} by using a reasoning augmentation model.
Given a user input speech, we first use an ASR model to transcribe the audio and feed the transcription into the reasoning augmentation model to generate a response, which will be used as the reasoning for \textsc{Stitch-R}.
Next, we crop the reasoning into chunks of $N'_{token}$ enclosed by \texttt{[SOPR]} and \texttt{[EOPR]}, and then feed the speech input and the first reasoning chunk to \textsc{Stitch-R} and generate the first text-speech interleaved token span.
After the first text-speech interleaved tokens are generated, we append the second reasoning chunk to the input and generate the second text-speech interleaved token spans.
We continue the above process to generate the full response.

In the following experiment, we replace the text reasoning of \textsc{Stitch-R} using four models: GLM-4-9B-Chat~\citep{glm2024chatglm}, Llama-3.1-8B, Llama-3.2-3B, and Llama-3.2-1B~\citep{grattafiori2024llama}.
The results are shown in Figure~\ref{fig:Images/augment_reasoning.pdf} in the Appendix, and we summarize the main findings below.
In general, stronger reasoning augmentation models (GLM-4-9B-Chat and larger Llama models) can improve the accuracy of the spoken response, while weaker reasoning augmentation model like Llama-3.2-1B worsens the performance of \textsc{Stitch-R}.
This shows that \textsc{Stitch-R} can incorporate the reasoning ability of stronger models to improve the spoken answer.
The experiment also shows that \textsc{Stitch-R} really considers the contents in the reasoning to form the spoken response, instead of independently generating the reasoning and not using that reasoning in the spoken responses.

\section{Conclusion}
\label{section: conclusion}

In this paper, we introduce \textsc{Stitch}, a generation framework for SLM that enables simultaneous thinking and speaking.
We achieve simultaneous thinking and speaking by using the long duration of an audio chunk to generate additional reasoning tokens before generating the speech tokens for the next audio chunk.
By design, \textsc{Stitch-S} induces no additional latency compared to the models that cannot generate reasoning before they speak.
On five math reasoning QA datasets, we show that \textsc{Stitch-R} and \textsc{Stitch-S} significantly improve the accuracy over non-reasoning baselines by over 15\%.
On non-reasoning datasets, \textsc{Stitch-R} and \textsc{Stitch-S} perform comparably well with non-reasoning baselines.
Overall, the low latency and high performance of \textsc{Stitch} show a promising future direction for spoken language models.

\section{Ethics Statement}
\label{Ethics}
Our paper fully follows the ICLR Code of Ethics.
We do not see any potential harm in the paper.

\bibliography{iclr2025_conference}
\bibliographystyle{iclr2025_conference}

\appendix

\section{\rev{Change Log}}

The following change log is included to help readers understand the modifications made throughout the author-reviewer discussion period.
\paragraph{Sep. 24th}
The initial submission version.

\paragraph{Nov. 19th}
First revision based on the initial reviews from the reviewer.
The modifications are highlighted in \rev{blue}.
The main differences are summarized as follows:
\begin{enumerate}
    \item \textbf{Explain the difficulty of fine-tuning Qwen-2.5-Omni}: In Section~\ref{section: Method},  we add a sentence to explain why we do not fine-tune Qwen-2.5-Omni.
    \item \textbf{Statistical tests in Section~\ref{section: main results}}: We run McNemar's test to see if the performance between no-reasoning baselines and other models is statistically significant and report the results in Section~\ref{section: main results}.
    \item \textbf{Explaining that reasoning always finishes before text-speech segments end}: In Section~\ref{section: main results}, we include a paragraph explaining that the reasoning segments generated by STITCH will always be less than the text/speech segments.
    \item \textbf{Human evaluation in Section~\ref{section: main results}}: We add a paragraph in Section~\ref{section: main results}, where we run a human evaluation comparing the responsiveness of different pairs of models.
    \item \textbf{Explain when to use STITCH-R/STITCH-S}: Based on all the previous experiment results, in the last paragraph in Section~\ref{section: main results}, we explain when to use STITCH-R and STITCH-S.
    \item \textbf{Qualitative results moved to Section~\ref{subsection: Qualitative Results}}: We move the qualitative results into the main content, where we illustrate how the reasoning of STITCH-R and STITCH-S precedes the text responses most of the time.
    \item \textbf{Add Figure~\ref{fig:Different t syn}}: We add Figure~\ref{fig:Different t syn} to show that different $t_{\rm syn}$ will not break the thinking while speaking pattern of STITCH as long as $t_{\rm syn} < t_{\rm chunk}$.
\end{enumerate}

\begin{figure}[ht!]
    \centering
    \begin{subfigure}[b]{0.32\textwidth}
        \includegraphics[width = \textwidth]{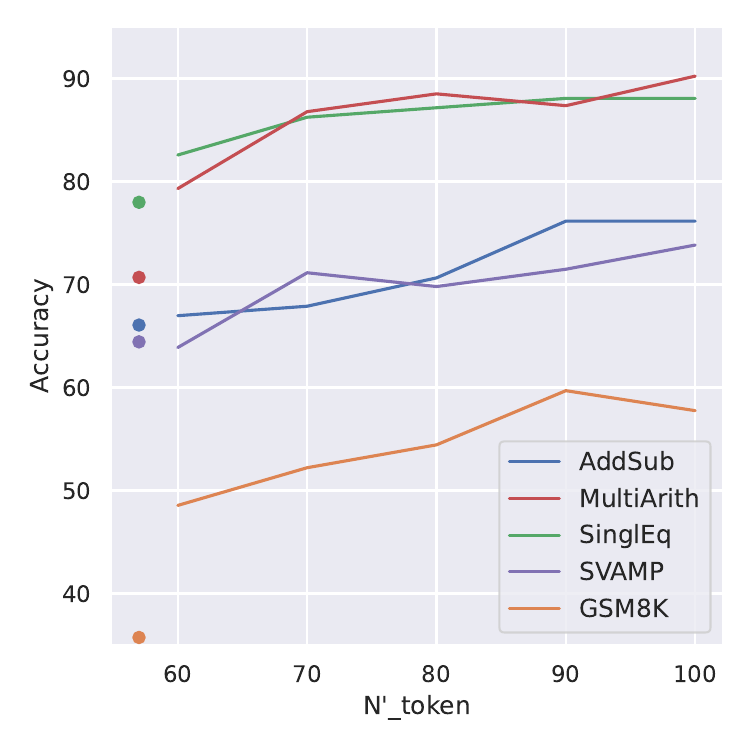}
        \caption{}
        \label{fig:Images/T2S2-I.pdf}
    \end{subfigure}
    \hfill
    \begin{subfigure}[b]{0.32\textwidth}
        \includegraphics[width = \textwidth]{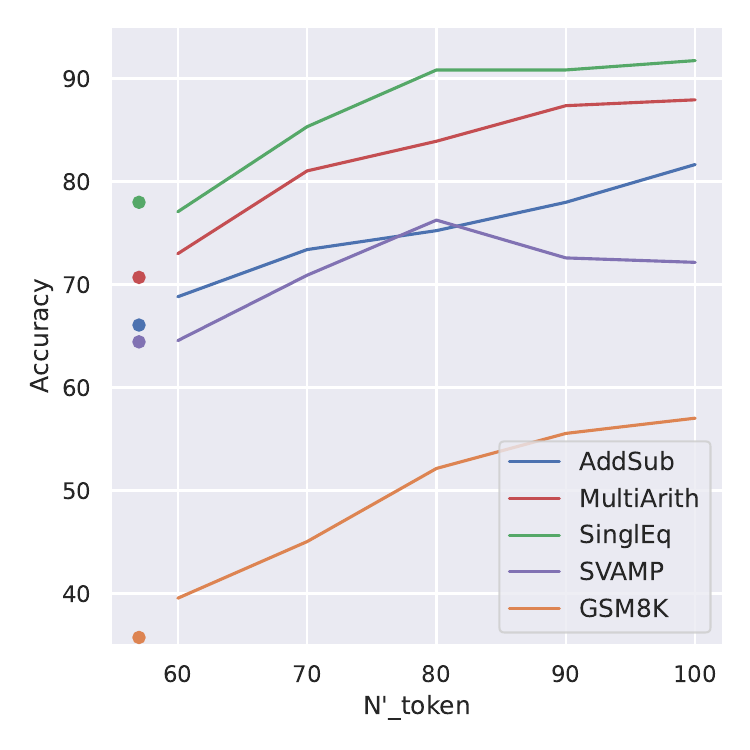}
        \caption{}
        \label{fig:Images/T2S2-RI.pdf}
    \end{subfigure}
    \hfill
    \begin{subfigure}[b]{0.32\textwidth}
        \includegraphics[width = \textwidth]{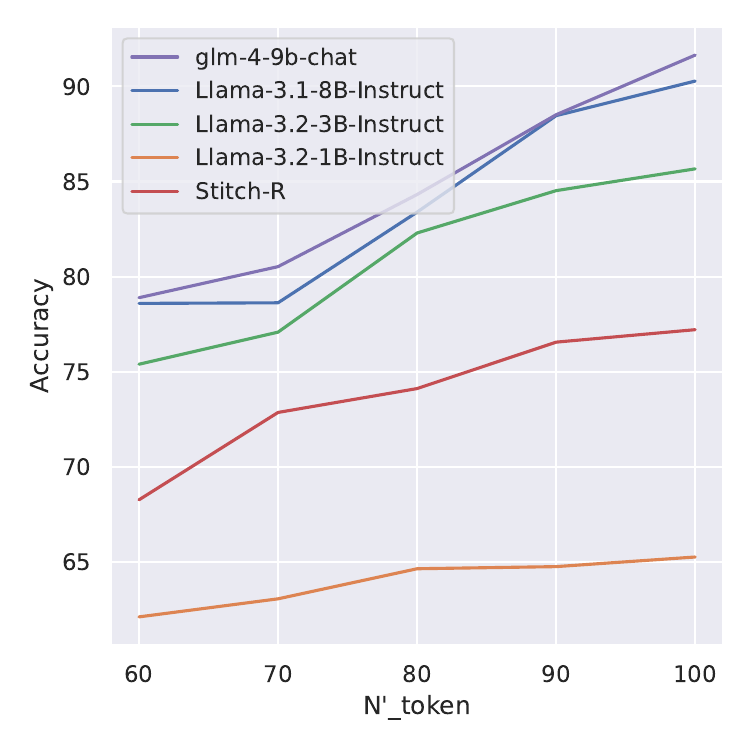}
        \caption{}
        \label{fig:Images/augment_reasoning.pdf}
    \end{subfigure}
    \caption{
    Figure~\ref{fig:Images/T2S2-I.pdf} and \ref{fig:Images/T2S2-RI.pdf} show the accuracy when varying $N'_{token}$ for \textsc{Stitch-R} and \textsc{Stitch-S}, respectively, and the dots in the figure are the performance of the "\textit{no reasoning}" baseline (Section~\ref{subsection: Adjust Length}).
    Figure~\ref{fig:Images/augment_reasoning.pdf} shows the performance when using a reasoning augmentation model to generate the text reasoning for \textsc{Stitch-R} (Section~\ref{subsection: Using text reasoning from Other Models}); the accuracy is averaged over five math reasoning datasets.} 
    \label{fig:analysis}
\end{figure}

\paragraph{Feb. 8th}

Camera-ready version. 
Minor edits for typos and add links for datasets and demos.

\section{Dataset}
\label{Appendix: Section: Dataset}

\subsection{Training Dataset}
\label{appendix: subsection: Training Dataset}
We use three types of training data: dialogue datasets, math reasoning datasets, and knowledge-intensive QA datasets.
The sizes of the datasets are in Table~\ref{tab:training-data}.
We explain how we construct our training datasets from those existing datasets as follows.

\paragraph{Dialogue Datasets}
VoiceAssistant400K~\citep{xiezhifei2024miniomni} is a spoken dialogue dataset built from a text dialogue dataset, including OpenAssistant~\citep{2023openassistant} and HH-RLHF~\citep{bai2022training}.
We subsample the dataset by filtering out (1) the instances whose \texttt{`round`} attribute is None or 0, (2) the instances from the subset \texttt{identity}, and (3) the instances whose response audio is less than 2 seconds.
Given a user query and the response or answer from the dataset, we use GPT-4o to generate a text reasoning given some in-context samples.
GPT-4o is instructed to generate CoTs that understand the user's intention, recall relevant information, and provide the final answer.
The prompt we use is shown in Table~\ref{tab:voice assistant 400K prompt}.

\paragraph{Math Reasoning Datasets}
We include three math reasoning datasets for fine-tuning: Tulu-3-Persona Math, Tulu-3-Persona-Math-Grade, and Tulu-3-Persona-Algebra.
The datasets include the question and the CoT answer in text.
We directly use the ground truth CoT answer as the text reasoning.
To obtain the spoken answer, we prompt GPT-4o to rewrite the answer such that the answer is suitable for speech communication and then synthesize it with GPT-4o-mini-TTS.
The voice is sampled from the following list: \textit{alloy, ash, ballad, coral, echo}.
The prompt used to rewrite the spoken answer is shown in Table~\ref{tab: spoken answer rewrite}.
The question audio is also synthesized with GPT-4o-mini-TTS, while the list of voices we sampled from is the same as the previous.
The voice of the spoken question and response is selected randomly and independently.
The total size of the math reasoning datasets used for fine-tuning is 220K.

\paragraph{Knowledge-Intensive Datasets}
We include two knowledge-intensive QA datasets, Natural Question (NQ)~\citep{kwiatkowski-etal-2019-natural} and TriviaQA (TQA)~\citep{joshi-etal-2017-triviaqa}, to preserve the SLM's ability to respond to knowledge-seeking questions.
These two datasets are text-based datasets.
To obtain the dataset for fine-tuning, we rewrite the ground truth answer from the dataset to make it more suitable for spoken communication, and then synthesize the question and answer audio using GPT-4o-mini-TTS.
We sample from the same subset of voices randomly.
The text reasoning is also generated by GPT-4o by prompting it with the question and ground truth answer, and asking it to generate an inner monologue that understands the user's intention, recalls relevant information, and provides the final answer.
The prompts used for rewriting the answer and generating the reasoning are shown in Table~\ref{tab:spoken-answer-factual} and Table~\ref{tab:factual-inner-thinking}, respectively.
When subsampling from NQ and TQA, we simply take the first 50K instances from NQ and the first 20K instances from TQA without shuffling the dataset before subsampling.
The size of knowledge-intensive datasets is around 70K instances.

\begin{table}[ht!]
    \centering
    \begin{adjustbox}{max width=\textwidth}
        \begin{tabular}{lcc}
            \toprule
            Dataset          & Size in Training Data & Download Link \\
            \midrule
            VoiceAssistant   & 177K                  & \href{https://huggingface.co/datasets/gpt-omni/VoiceAssistant-400K}{gpt-omni/VoiceAssistant-400K} \\
            Tulu3 Math       & 150K                  & \href{https://huggingface.co/datasets/allenai/tulu-3-sft-personas-math}{allenai/tulu-3-sft-personas-math} \\
            Tulu3 Algebra    & 20K                   & \href{https://huggingface.co/datasets/allenai/tulu-3-sft-personas-algebra}{allenai/tulu-3-sft-personas-algebra} \\
            Tulu3 GSM        & 50K                   & \href{https://huggingface.co/datasets/allenai/tulu-3-sft-personas-math-grade}{allenai/tulu-3-sft-personas-math-grade} \\
            Natural Question & 50K                   & \href{https://huggingface.co/datasets/google-research-datasets/natural_questions}{google-research-datasets/natural\_questions} \\
            TriviaQA         & 20K                   & \href{https://huggingface.co/datasets/mandarjoshi/trivia_qa}{mandarjoshi/trivia\_qa} \\
            \bottomrule
        \end{tabular}
    \end{adjustbox}
    \caption{Training data sizes for each dataset.
    The download link in the above table is the link where we download the original datasets and then apply our filtering and preprocessing.
    Directly downloading from those links will not get the training dataset we use. 
    Please follow the steps detailed in Appendix~\ref{appendix: subsection: Training Dataset} to construct the training data.}
    \label{tab:training-data}
\end{table}

\subsection{Testing Dataset}
\label{Appendix: subsection: Testing Dataset}
The testing dataset statistics are shown in Table~\ref{tab:test datasets}.
We release the audio of the math datasets we use at \url{https://huggingface.co/datasets/dcml0714/speech_math}.
We use \href{https://github.com/MoonshotAI/Kimi-Audio-Evalkit}{Kimi-Audio-Evalkit}~\citep{ding2025kimi} to download OpenAudioBench.
All our evaluation is run using Kimi-Audio-Evalkit.
When evaluating the accuracy for math reasoning datasets and knowledge-intensive QA datasets, we use GPT-4o to judge if the model prediction matches the ground truth; this is supported by Kimi-Audio-Evalkit.
The numbers of our result for GLM-4-Voice slightly differ from the numbers reported in \citet{ding2025kimi} since we use GPT-4o as the judge, while \citet{ding2025kimi} use GPT-4o-mini as the judge.
In the original paper that proposes OpenAudioBench, \citet{baichuanaudio} also use GPT-4o as the judge, and our numbers for GLM-4-Voice are very close to those reported in \citet{baichuanaudio}.

\begin{table}[h!]
    \centering

    \begin{subtable}[t]{0.8\textwidth}
        \centering
        \begin{adjustbox}{max width=\textwidth}
            \begin{tabular}{ccccc}
                \toprule
                \multicolumn{5}{c}{Math Reasoning Datasets} \\
                \midrule
                AddSub & MultiArith & SinglEq & SVAMP & GSM8K \\
                109    & 174        & 109     & 299   & 1310  \\
                \bottomrule
            \end{tabular}
        \end{adjustbox}
        \subcaption{}
        \label{tab:math-reasoning-dataset}
    \end{subtable}

    \vspace{1em}

    \begin{subtable}[t]{0.8\textwidth}
        \centering
        \begin{adjustbox}{max width=\textwidth}
            \begin{tabular}{cccc}
                \toprule
                \multicolumn{4}{c}{OpenAudioBench} \\
                \midrule
                Llama Questions & TriviaQA & Web Questions & AlpacaEval \\
                300             & 1000     & 1000         & 199         \\
                \bottomrule
            \end{tabular}
        \end{adjustbox}
        \subcaption{}
        \label{tab:openaudiobench}
    \end{subtable}

    \caption{Counts for Math Reasoning benchmarks (top) and OpenAudioBench (bottom).}
    \label{tab:test datasets}
\end{table}

\begin{table}[t!]
    \scriptsize 
    \centering
    \begin{tabularx}{\textwidth}{X}
        \toprule
        \bg{question}{\textbf{Question}:} Josh decides to try flipping a house.  He buys a house for \$80,000 and then puts in \$50,000 in repairs.  This increased the value of the house by 150\%.  How much profit did he make?\\
        \bg{question}{\textbf{Response}:}\\
        \bgp{reason}{
            \texttt{[SOPR]}
            \#\# Inner Monologue\quad
            This is a math question. I need to think step by step. To solve this problem, we need to determine the final value of the house after the repairs and then calculate the profit Josh made. \\
            1. **Initial Purchase Cost**: Josh buys the house for \$80,000.\\
            2. **Cost of Repairs**: He spends \$50,000 on repairs.\\
            3. **Total Investment**: The total investment is the sum of the purchase cost and the repair cost
            \texttt{[EOPR]}    
        }\\
        \bgp{trans}{Josh spent \$80,000 on the house and \$50,}
        \bgp{speech}{\texttt{<audio\_tokens>} (...)} \\
        \bgp{reason}{
            \texttt{[SOPR]}
            Total Investment = \$80,000 + \$50,000 = \$130,000 \\
            4. **Increase in Value**: The value of the house increases by 150\% after the repairs. To find the new value of the house, we calculate 150\% of the initial purchase cost and add it to the initial purchase cost. \quad Increase in Value = 1.5 $\times$ \$80,
            \texttt{[EOPR]}    
        }\\
        \bgp{trans}{000 on repairs, totaling \$130,000. After repairs,}
        \bgp{speech}{\texttt{<audio\_tokens>} (...)} \\
        \bgp{reason}{
            \texttt{[SOPR]} \quad
            000 = \$120,000. New Value of the House = \$80,000 + \$120,000 = \$200,000 \\
            5. **Profit Calculation**: The profit is the difference between the new value of the house and the total investment.\quad Profit = \$200,000 - \$130,000 = \$70,000 \\
            Thus, Josh made a profit of    \texttt{[EOPR]}    
        }\\
        \bgp{trans}{the house's value increased by 150\%, making it \$200}
        \bgp{speech}{\texttt{<audio\_tokens>} (...)} \\
        \bgp{reason}{
            \texttt{[SOPR]}\quad \$70,000.\quad \#\#\#\# \$70,000
            \texttt{[EOPR][EOR]}    
        }\\
        \bgp{trans}{,000. So, his profit is \$200,000 minus}
        \bgp{speech}{\texttt{<audio\_tokens>} (...)} 
        \bgp{trans}{\$130,000, which equals \$70,000. The}
        \bgp{speech}{\texttt{<audio\_tokens>} (...)} 
        \bgp{trans}{ answer is 70,000.}
        \bgp{speech}{\texttt{<audio\_tokens>} (...)} \\
        \midrule
        \bg{question}{\textbf{Question}:} Hannah's city is having a big display of fireworks for the 4th of July. They're going to set off 15 boxes of 20 fireworks each. Hannah's house is at the right angle to see 40\% of the city's fireworks. Hannah will also set off 3 boxes of 5 fireworks each in her backyard. How many fireworks will Hannah see in total? \\
        \bg{question}{\textbf{Response}:}\\
        \bgp{reason}{\texttt{[SOPR]} \quad
        \#\# Inner Monologue        
                This is a math question. I need to think step by step. To solve this problem, we need to calculate the total number of fireworks Hannah will see from both the city’s display and her own backyard display.
                
                **Step 1: Calculate the total number of fireworks from the city’s display.**
                
                The city’s display consists of 15 boxes of 20 fireworks each. To find the total number of fireworks from the city’s display, we multiply the number of \texttt{[EOPR]}} 
        \bgp{trans}{Hannah’s house will see 40\% of the city’s}  \bgp{speech}{\texttt{<audio\_tokens>} (...)} 
        
        \bgp{reason}{\texttt{[SOPR]} boxes by the number of fireworks in each box: 
        \quad
        15 \text{ boxes} $\times$ 20 \text{ fireworks per box} = 300 \text{ fireworks}
        
                **Step 2: Calculate the total number of fireworks from Hannah’s backyard display.**
                
                Hannah sets off 3 boxes of 5 fireworks each in her backyard. To find the total number of fireworks from her backyard display, we multiply the number of boxes by the number of fireworks in each box:
                \texttt{[EOPR]}}
        \bgp{trans}{fireworks, which is 120. Plus, she has 3} \bgp{speech}{\texttt{<audio\_tokens>} (...)} \\
        \bgp{reason}{\texttt{[SOPR]}
        \quad
        3 \text{ boxes} $\times$ 5 \text{ fireworks per box} = 15 \text{ fireworks}\\
                **Step 3: Calculate the total number of fireworks Hannah will see.**
                
                Hannah will see 40\% of the city’s fireworks. To find out how many fireworks this is, we calculate 40\% of the total fireworks from the city’s display:\quad 
                    40\% \text{ of } 300 \text{ fireworks} = \texttt{[EOPR]}} 
         \bgp{trans}{boxes of her own fireworks, adding 15 more. So in} \bgp{speech}{\texttt{<audio\_tokens>} (...)} \\
         \bgp{reason}{ \texttt{[SOPR]}
         \quad 0.40 $\times$ 300 = 120 fireworks\\
         **Step 4: Add the fireworks from the city’s display and Hannah’s backyard display.**
        Finally, we add the fireworks from the city’s display and Hannah’s backyard display to find the total number of fireworks Hannah will see:
        120 \text{ fireworks from city display} + 15 \text{ fireworks from backyard display} = 135 \text{ fireworks}\quad
                \#\#\#\# 135 \texttt{\texttt{[EOPR]}} \texttt{[EOR]}} 
         \bgp{trans}{total, she’ll see 135 fireworks.} \bgp{speech}{\texttt{<audio\_tokens>} (...)} \\
        \bottomrule
    \end{tabularx}
    \caption{Example outputs from \textsc{Stitch-R}.
    The text reasoning tokens are highlighted in \bg{reason}{green}; the text tokens are highlighted in \bg{trans}{orange}; the audio tokens, which are omitted for simplicity, are highlighted in \bg{speech}{red}.
    We remove and insert some newlines for formatting.}
    \label{tab:T2S2-I samples}
\end{table}

\section{Training Details}
\label{Appendix: section: Training Details}

We use LlamaFactory~\citep{zheng-etal-2024-llamafactory} to fine-tune the model.
Fine-tuning a model using LlamaFactory~\citep{zheng-etal-2024-llamafactory} takes 17 hours on 32 A100-80GB GPUs.
To train with LlamaFactory, we use the speech tokenizer to encode the audio of the question and answer into speech tokens.
Since LlamaFactory does not support fine-tuning GLM-4-Voice, the reader will need to slightly modify some source code in LlamaFactory to allow it to support GLM-4-Voice.
We provide the exact \texttt{yaml} configuration file for fine-tuning below, including the detailed list of hyperparameters we use.

\lstset{
    language=yaml,
    basicstyle=\ttfamily\footnotesize,
    breaklines=true,
    frame=single,
    numbers=left,
    numberstyle=\tiny\color{gray},
    numbersep=5pt,
    showstringspaces=false
}

\begin{lstlisting}[language=yaml]
### model
model_name_or_path: THUDM/glm-4-voice-9b
trust_remote_code: true

### method
stage: sft
do_train: true
finetuning_type: full

### dataset
template: glm4voice
cutoff_len: 8192
overwrite_cache: true
preprocessing_num_workers: 16
dataloader_num_workers: 4
cache_dir: ./

### output
logging_steps: 10
save_steps: 500
plot_loss: true
overwrite_output_dir: true
save_only_model: false
report_to: none  

### train
per_device_train_batch_size: 2
gradient_accumulation_steps: 8
learning_rate: 1.0e-5
num_train_epochs: 2.0
lr_scheduler_type: cosine
warmup_ratio: 0.1
bf16: true
ddp_timeout: 180000000
enable_liger_kernel: true
deepspeed: examples/deepspeed/ds_z2_config.json 
\end{lstlisting}

\section{\rev{Human Evaluation}}
\label{Appendix: Human Evaluation}

\rev{
This section elaborates on how we conduct human evaluations to evaluate the responsiveness of different model pairs.
We conduct pairwise evaluation of the responsiveness of the following model pairs:
(1) STITCH-R vs TBS, (2) STITCH-S vs TBS, (3) STITCH-R vs STITCH-S, and (4) STITCH-S vs no-reasoning.
We select the above model pairs because we believe they can form a partial order of the responsiveness of the four models.}

\rev{
We generate the responses of each model from the same 100 questions from GSM8K; the reason for selecting GSM8K is that the number of reasoning tokens required for this dataset is longer, as shown in Table~\ref{tab:merged-comparison}, which may make the difference of responsiveness among the models more observable.
For each audio, we insert a silence to represent the first packet latency.
The first packet latency considers the prefilling and decoding time for the SLM, and the speech synthesis latency of the speech decoder.
The prefilling and decoding latency is profiled using an A100 GPU with vLLM, while the speech synthesis latency is calculated based on the reported latency from Cosyvoice.}

\rev{The human evaluators are given a question in GSM8K and two audios from a pair of models, and they are asked to judge the responsiveness of the two models on a 7-point comparison category rating, following the P.808~\citep{rec2018p808}, using the following rubrics:}
\rev{\begin{itemize}
        \item[+3]: A is more responsive than B
        \item[+2]: A is somewhat responsive than B
        \item[+1]: A is slightly responsive than B
        \item[0]: A and B are equally responsive
        \item[-1]: B is slightly responsive than A
        \item[-2]: B is somewhat responsive than A
        \item[-3]: B is more responsive than A
\end{itemize}}

\begin{figure}[t!]
\includegraphics[clip, trim = 0px 0px 0px 0px, width = \textwidth]{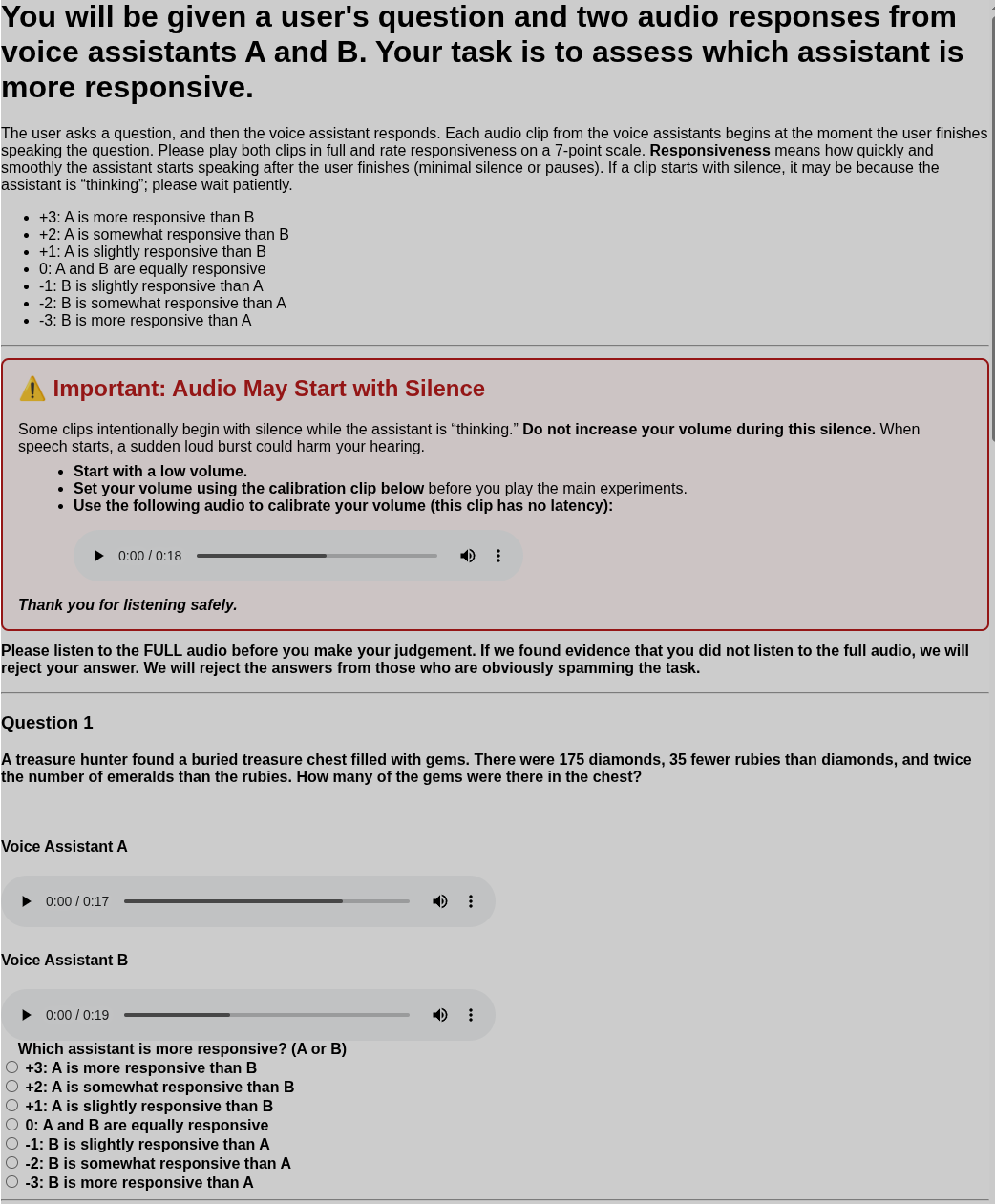}
\caption{\rev{The interface of the human evaluation.}} 
    \label{fig:human eval}
\end{figure}

\begin{figure}[ht]
    \centering
    \begin{subfigure}{0.95\textwidth}
        \centering
        \includegraphics[width=\linewidth]{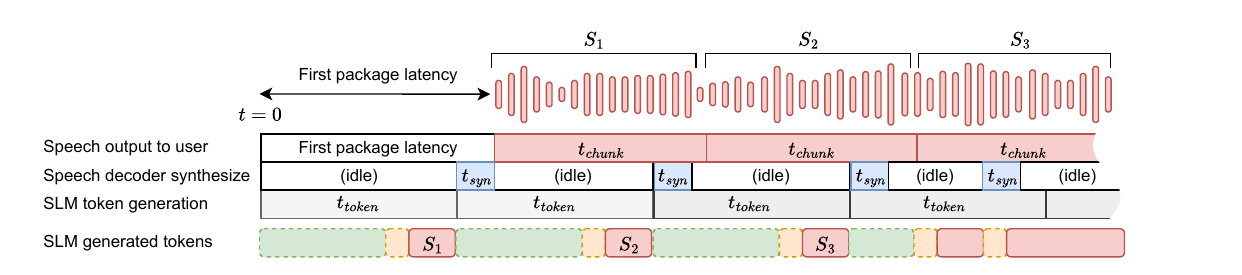}
        \caption{An illustration of timing diagram with a smaller $t_{syn}$, the time for synthesizing $N_{\rm speech}$ tokens in the the corresponding audio.} 
        \label{fig:image1}
    \end{subfigure}
    \hfill
    \begin{subfigure}{0.95\textwidth}
        \centering
        \includegraphics[width=\linewidth]{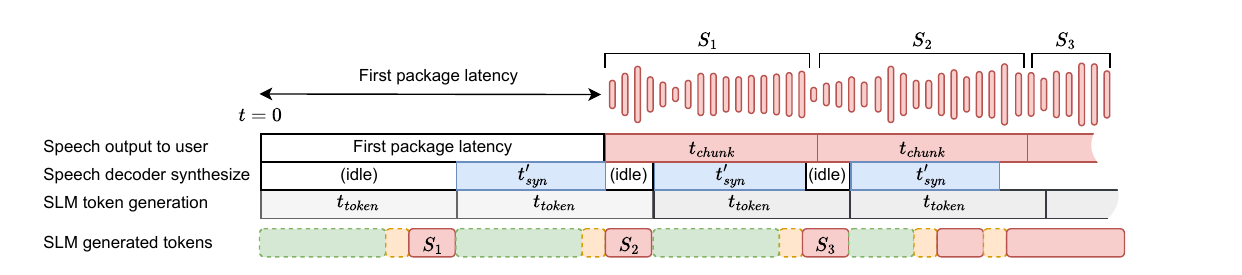}
        \caption{An illustration of timing diagram with a larger $t'_{syn}$, the time for synthesizing $N_{\rm speech}$ tokens in the the corresponding audio.}
        \label{fig:image2}
    \end{subfigure}
    \caption{\rev{The above two figures show the same STITCH-R model with different $t_{\rm syn}$.
    The upper one has a smaller $t_{\rm syn}$, while the lower one has a larger $t_{\rm syn}$.
    The figure is used to show that the speech decoder and the token generation of STITCH can be run in parallel.
    Consequently, as long as the $t_{\rm syn}<t_{\rm chunk}$, two audio chunks can be played seamlessly to the user.
    The difference in $t_{\rm syn}$ only affects the first packet latency.}}
    \label{fig:Different t syn}
\end{figure}

\rev{We use Amazon MTurk (AMT) to recruit the human evaluators, and each pair of comparisons is rated by three annotators.
We require the annotators to have a HIT approval rate $\geq$ 98 and to hold a master qualification on AMT.
We also only include participants from the US, UK, Australia, and New Zealand in this study.
The interface for the human evaluation is shown in Figure~\ref{fig:human eval}.
Human evaluators are paid \$1.70 for rating 8 pairs of responses, and we estimate that the hourly rate is roughly \$12.75.}

\rev{The results are summarized in Table~\ref{tab: human evaluation}.
STITCH-S reaches a score of 1.687 against TBS, while STITCH-R obtains 1.164 compared to TBS, indicating that while both STITCH models are more responsive than TBS, the responsiveness of STITCH-S is slightly better than STITCH-R.
The pairwise comparison between STITCH-R and STITCH-S also shows that STITCH-S is perceived as more responsive than STITCH-R.
Comparing STITCH-S and the no-reasoning baseline, the average score is 0.290, while the 95\% confidence interval is $[0.131,0.449]$, meaning that the difference in the responsiveness is not very significant.}

\begin{table}
    \centering
    \begin{adjustbox}{max width=\textwidth}
        \begin{tabular}{lcccc}
            \toprule
            Model A & STITCH-R & STITCH-S  & STITCH-R  & STITCH-S  \\
            & vs & vs & vs & vs \\
            Model B & TBS & TBS & STITCH-S & no-reasoning \\
            \midrule
            Mean score & 1.164 & 1.687 & -0.850 & 0.290 \\
            \bottomrule
        \end{tabular}
    \end{adjustbox}
    \caption{\rev{The mean responsiveness scores of the four pairs of models.
    A positive score indicates that the upper model (model A) is more responsive than the lower one (model B), while a negative number means the lower model is more responsive.
    The larger the absolute value of the score is, the more obvious the responsiveness of the two models is for human evaluators.}}
    \label{tab: human evaluation}
\end{table}

\section{Detailed Latency Calculation}
\label{appendix: Detailed Latency Calculation}
In the main content, we assume that the duration of each audio chunk $t_{\text{chunk}}$ synthesized from $N_{\text{speech}}$ tokens is the same, and we use $t_{\text{chunk}}=2$ seconds in the paper.
This is a simplification to make the discussion easier.
In fact, for the first chunk of 26 speech tokens, the corresponding audio is about 1.6 seconds, while the later audio chunks all have a duration of about 2 seconds.
As a result, a safer $N_{\text{reason}}$ on an A100-80G GPU, with token per second around 86, should be $86 \times 1.6 - (13+26)=98.6$, which is still around 100, the number we use in our paper.
It is also possible to make only the first reasoning chunk shorter while all the later reasoning chunks have $N_{\text{reason}}=100$.
If the model is deployed on H100-80G GPU, the token per second can be as high as 123, the maximum number for $N_{\text{token}}$ can be $123 \times 1.6 - (13+26)=157.8$.
Using speculative decoding~\citep{leviathan2023fast} to speed up and increase the number of reasoning tokens for each chunk is also a valid option.

\section{Unsuccessful Attempts}
\label{appendix: Unsuccessful Attempts}

We detail the unsuccessful attempts and negative findings observed in our experiments. Our aim is to guide future research by highlighting these pitfalls so that others can avoid them.

\paragraph{LoRA Fine-tuning}
Initially, we tried to use LoRA~\citep{hu2022lora} for fine-tuning.
However, when fine-tuning TBS using LoRA, we found that the resulting model cannot generate useful reasoning.
After fine-tuning, the performance on GSM8K is still around 35\%, which is almost the same as the accuracy of GLM-4-Voice.
By observing the outputs from the fine-tuned model, we observed that the model can generate outputs that satisfy the output format in the training data, i.e., reasoning and then interleaved text-speech tokens.
However, the reasoning is seldom correct and can make very trivial mistakes.
We suspect that this is because teaching GLM-4-Voice mathematical reasoning requires significant model parameter updates, making LoRA fine-tuning a bad choice.

\paragraph{Fine-tuning only on Math Datasets}
We have tried to fine-tune only on the math datasets to teach the model reasoning. 
However, we found that the resulting model will treat every input query as a math question and list some equations to solve the question, even if the question simply asks "\textit{Which city is located at the intersection of the Tigris and Euphrates rivers?}"
This is possibly because the model overfits to the pattern in the training data and forgets how to solve other non-math questions.
However, this can be easily solved by including diverse types of tasks in the training data, which is what we do in the end.

\paragraph{Customized Attention Mask}
When training the \textsc{Stitch-R} model, we tried with a variant with customized attention masks.
Precisely, we make the reasoning tokens unable to attend to the text tokens and speech tokens, while the text and speech tokens can attend to all types of tokens.
The attention mask still remains causal.
The rationale for this customized attention mask is to make the reasoning tokens feel like there is no \textit{gap} between two chunks of reasoning tokens, since they cannot see the text and speech tokens that separate two chunks of reasoning tokens.
We use the customized attention mask during training and inference.
However, we found that the results on the math reasoning datasets are similar to the results of not using a customized attention mask.
Considering that the customized attention mask requires some customization during training, complicates the inference procedure, and does not improve the performance, we do not recommend this configuration.

\begin{table}[t!]
    \centering

        \begin{adjustbox}{max width=\textwidth}
            \begin{tabular}{cccc|ccccc|c}
                \toprule
                
                 \multirow{2}{*}{Id}&\multirow{2}{*}{Config}   
                   & \multicolumn{2}{c}{\textit{Use Reasoning}}  
                  & \multicolumn{5}{|c|}{\textit{Datasets}} 
                  & \multirow{2}{*}{\textit{Avg.}} \\
                   
                   &                   & Train  
                  & Infer.  
                  & \textit{AddSub} 
                  & \textit{MultiArith} 
                  & \textit{SinglEq} 
                  & \textit{SVAMP} 
                  & \textit{GSM8K} 
                  &                     \\
                \midrule
                 &\multicolumn{8}{c}{\textit{\textbf{Baselines}}} &  \\
                \midrule
                 1&GLM-4-Voice             
                   & –       
                  & \ding{56} 
                  & 49.54& 55.75& 66.06& 40.13& 25.80& 47.46\\
                 2&No reasoning            
                   &\ding{56}
                  & \ding{56}
                  & 62.39& 65.52& 71.56& 57.53& 33.13& 58.03\\
                  3&Mix reasoning
                   & \ding{52}+\ding{56}
                  & \ding{56}
                  & 62.39& 77.59& 85.32& 60.87& 36.49& 64.53\\
                  \midrule
                   &\multicolumn{8}{c}{\textit{\textbf{Full Reasoning before Speaking}}} &  \\
                  \midrule
                  4&TBS                  
                   & \ding{52}
                  & \ding{52}
                  & 70.64& 79.31& 84.40& 71.57& 60.69& 73.32\\
                
                  5&Mix reasoning
                   & \ding{52}+\ding{56}         
                  & \ding{52}
                  & 67.89& 86.21& 83.49& 69.57& 59.92& 73.42\\
                \midrule
                 &\multicolumn{8}{c}{\textit{\textbf{Proposed Methods: Simultaneous Thinking and Speaking}}} &  \\
                \midrule

                 6&\textsc{Stitch-R}                  
                   & \ding{52}
                  & \ding{52}
                  & 67.89& 83.33& 88.07& 69.13& 53.05& 72.29\\
                 7&\textsc{Stitch-S}                
                   & \ding{52}
                  & \ding{52}
                  & 73.39& 84.48& 81.65& 66.78& 52.18& 71.70\\
                \bottomrule
            \end{tabular}
        \end{adjustbox}

    \caption{The performance on math reasoning datasets when evaluating the transcription of the speech output.
    We use Whisper-large-v3~\cite{radford2023robust} to transcribe the audio. 
    The \textit{"use reasoning"} columns denote whether text reasoning is in the training data or is generated during inference.}
    \label{tab:reasoning asr}
\end{table}

\begin{table}[t!]
    \scriptsize 
    \centering
    \begin{tabularx}{\textwidth}{X}
        \toprule
        \bg{question}{\textbf{Question}:} Claire makes a 3 egg omelet every morning for breakfast.  How many dozens of eggs will she eat in 4 weeks?\\
        \bg{question}{\textbf{Response}:}\\
        \bgp{trans}{Claire makes a 3-egg omelette every morning}
        \bgp{speech}{\texttt{<audio\_tokens>} (...)} \\
        \bgp{reason}{
            \texttt{[SOPR]}
            \#\# Inner Monologue\quad
            This is a math question. I need to think step by step. To solve this problem, we need to determine how many dozens of eggs Claire will eat in 4 weeks if she makes a 3-egg omelette every morning.\\
            1. **Calculate the number of days in 4 weeks:**\\
            - There are 7 days in a week.   \\
            - Therefore, in 4 weeks, there are (4 $\times$ 7 = 28
            \texttt{[EOPR]}    
        }\\
        \bgp{trans}{, and there are 28 days in 4 weeks. That}
        \bgp{speech}{\texttt{<audio\_tokens>} (...)} \\
        \bgp{reason}{
            \texttt{[SOPR]} \quad
            
            ) days.\\
            2. **Determine the number of eggs Claire eats per day:**\\
            - Claire makes a 3-egg omelette every morning.\\
            - Therefore, she eats 3 eggs per day.\\
            3. **Calculate the total number of eggs Claire eats in 28 days:**\\
            - Since she eats 3 eggs per day, in 28 days, she will eat (3 $\times$ 28 = 84) eggs.\\
            4. **Convert the total number\texttt{[EOPR]}    
        }\\
        \bgp{trans}{'s 84 eggs in total. Since there are 12 eggs}
        \bgp{speech}{\texttt{<audio\_tokens>} (...)} \\
        \bgp{reason}{
            \texttt{[SOPR]}\quad  of eggs from whole eggs to dozens:**\\
            - There are 12 eggs in a dozen.\\
            - To find out how many dozens of eggs she eats, divide the total number of eggs by 12.\\
            - ( \\frac\{84\}\{12\} = 7 ) dozens.\\
            Thus, Claire will eat 7 dozens of eggs in 4 weeks.\quad \#\#\#\# 7
            \texttt{[EOPR][EOR]}    
        }\\
        \bgp{trans}{in a dozen, she'll eat 7 dozens of eggs in}
        \bgp{speech}{\texttt{<audio\_tokens>} (...)} \\
        \bgp{trans}{ 4 weeks. The answer is 7 dozens.}
        \bgp{speech}{\texttt{<audio\_tokens>} (...)} \\
        \bottomrule
    \end{tabularx}
    \caption{An Example output from \textsc{Stitch-S}.
    The text reasoning tokens are highlighted in \bg{reason}{green}; the text tokens are highlighted in \bg{trans}{orange}; the audio tokens, which are omitted for simplicity, are highlighted in \bg{speech}{red}.
    We remove and insert some newlines for formatting.}
    \label{tab:T2S2-IR samples}
\end{table}

\begin{table}[ht]
    \centering
    \footnotesize
    \begin{adjustbox}{max width=\textwidth}
        \begin{tabular}{p{16cm}}
            \toprule
            \texttt{You are a helpful evaluation assistant. You will be given a question, its ground truth answer, and a prediction by an AI assistant. Your job is to evaluate the fluency of the answer into a single score. The score you can choose from is 1, 2, 3, 4, 5. Note that you **do not** need to judge the accuracy of the AI assistant’s prediction; you only need to consider how fluent the sentences are.}\\
            \\
            \texttt{\#\# Evaluation Criteria:}\\
            \texttt{Fluency and Coherence (1–5): This rating measures the quality of individual sentences—are they well‑written and grammatically correct? It also covers the overall organization and structure of the prediction.}\\
            \\
            \texttt{\#\# Output Format:}\\
            \texttt{Please provide some reasoning about how you evaluate the prediction and explain what factors affect your judgment. Conclude your answer with `Final verdict: score`, where `score` is an integer from [1,2,3,4,5].}\\
            \\
            \texttt{\#\# Question}\\
            \texttt{\{question\}}\\
            \\
            \texttt{\#\# Ground Truth answer}\\
            \texttt{\{answer\}}\\
            \\
            \texttt{\#\# Prediction by AI assistant}\\
            \texttt{\{prediction\}}\\
            \\
            \texttt{\#\# Evaluation Task}\\
            \texttt{How fluent and coherent is the prediction? (On a scale of 1–5, with 1 being the lowest)}\\
            \bottomrule
        \end{tabular}
    \end{adjustbox}
    \caption{The prompt used to evaluate the fluency of the text tokens from the model with LLM-as-a-judge.}
    \label{tab:fluency prompt}
\end{table}

\begin{table}[ht]
    \centering
    \footnotesize
    \begin{adjustbox}{max width=\textwidth}
        \begin{tabular}{p{16cm}}
            \toprule
            \texttt{You will be given a math question and its inner reasoning about the math question that is used to derive the final answer. Your job is to summarize the reasoning into a short and concise answer that is suitable to be spoken in dialogue. You should avoid using excessive equations and math notations but only briefly mention how the result is calculated. You don’t need to restate each step and intermediate results in the inner reasoning; just provide a brief overall summary.}\\
            \\
            \texttt{\#\# Example 1:}\\
            \texttt{\#\#\# Question \#\#\#}\\
            \texttt{Natalia sold clips to 48 of her friends in April, and then she sold half as many clips in May. How many clips did Natalia sell altogether in April and May?}\\
            \texttt{\#\#\# Inner reasoning \#\#\#}\\
            \texttt{I need to calculate the number of clips Natalia sold in April and May. I am given that she sold 48 in April, and I need to calculate how many she sold in May. We know that she sold half as many in May, so 48 / 2 = 24. She sold 24 in May. Combining the answer, she sold 48 + 24 = 72 in April and May.}\\
            \texttt{\#\#\# Summarized spoken answer \#\#\#}\\
            \texttt{In April she sold 48 clips and in May she sold half that amount, which is 24, so altogether she sold 72 clips.}\\
            \\
            \texttt{\#\# Example 2:}\\
            \texttt{\#\#\# Question \#\#\#}\\
            \texttt{Weng earns \$12 an hour for babysitting. Yesterday, she just did 50 minutes of babysitting. How much did she earn?}\\
            \texttt{\#\#\# Inner reasoning \#\#\#}\\
            \texttt{We need to calculate how much she earned in 50 minutes. We are given that she earns 12 per hour. 50 minutes corresponds to 50 / 60 = 5 / 6 hours. As a result, she can earn 12 * (5 / 6) = 10. The answer is 10.}\\
            \texttt{\#\#\# Summarized spoken answer \#\#\#}\\
            \texttt{She worked 50 minutes, which is five‑sixths of an hour at \$12 per hour, so she earned \$10.}\\
            \\
            \texttt{\#\# Example 3:}\\
            \texttt{\#\#\# Question \#\#\#}\\
            \texttt{Betty is saving money for a new wallet which costs \$100. Betty has only half of the money she needs. Her parents decided to give her \$15 for that purpose, and her grandparents twice as much as her parents. How much more money does Betty need to buy the wallet?}\\
            \texttt{\#\#\# Inner reasoning \#\#\#}\\
            \texttt{We need to calculate how much more money Betty needs. She has saved \$50, her parents gave \$15, and grandparents gave \$30, so she has \$95 in total and needs \$5 more. The answer is 5.}\\
            \texttt{\#\#\# Summarized spoken answer \#\#\#}\\
            \texttt{She has \$95 so far from her own savings and gifts, and since the wallet costs \$100, she still needs \$5.}\\
            \\
            \texttt{\# Now it is your turn}\\
            \texttt{\#\#\# Question \#\#\#}\\
            \texttt{\{question\}}\\
            \texttt{\#\#\# Inner reasoning \#\#\#}\\
            \texttt{\{reasoning\}}\\
            \texttt{\#\#\# Summarized spoken answer \#\#\#}\\
            \bottomrule
        \end{tabular}
    \end{adjustbox}
    \caption{The prompts used to rewrite the answers in the dataset into the spoken answer for the math reasoning dataset in the training data.
    The \texttt{\{question\}} is the placeholder of the question; the \texttt{\{reasoning\}} will be replaced by the reasoning, which is the original answer in the training data.}
    \label{tab: spoken answer rewrite}
\end{table}

\begin{table}[ht]
    \centering
    \footnotesize
    \begin{adjustbox}{max width=\textwidth}
        \begin{tabular}{p{16cm}}
            \toprule
            \texttt{You will be given a factual question and its answer. It is possible that there are multiple answers to the question, and we are only providing one for simplicity. Your job is to generate some inner thinking to derive the answer. Imagine that you are a human and you heard somebody ask you that question, and you should think before responding to the question in speech. Your inner thinking can include rephrasing the question into a better way for understanding, and clarifying any possible ambiguity, retrieving the correct answer and some relevant information. Your inner reasoning should include the short answer. You should not say "I know the answer" without explicitly saying the answer. You should not say "I need more information" and end your reasoning since this is not helpful at all.}\\
            \\
            \texttt{\#\# Example 1}\\
            \texttt{\#\# Question:}\\
            \texttt{What color was John Wilkes Booth's hair?}\\
            \texttt{\#\# Answer}\\
            \texttt{jet-black}\\
            \texttt{\#\# Inner thinking}\\
            \texttt{I think the user is asking about John Wilkes Booth who assassinated United States President Lincoln on April 14, 1865. His hair color is jet-black. Got the answer and ready to provide it to the user.}\\
            \\
            \texttt{\#\# Example 2}\\
            \texttt{\#\# Question:}\\
            \texttt{Can you make and receive calls in airplane mode?}\\
            \texttt{\#\# Answer}\\
            \texttt{No}\\
            \texttt{\#\# Inner thinking}\\
            \texttt{Airplane mode is a setting on smartphones and other portable devices that suspends all radio‑frequency transmissions. So you can't make or receive calls while it's on. The answer is no.}\\
            \\
            \texttt{\#\# Example 3}\\
            \texttt{\#\# Question:}\\
            \texttt{Who played "Will Munson" on "As the World Turns"?}\\
            \texttt{\#\# Answer}\\
            \texttt{Jesse Lee Soffer}\\
            \texttt{\#\# Inner thinking}\\
            \texttt{The user likely means the CBS soap "As the World Turns." The character Will Munson was most recently portrayed by Jesse Lee Soffer. Ready to speak the answer.}\\
            \\
            \texttt{\# Now it is your turn}\\
            \texttt{\#\# Question:}\\
            \texttt{\{question\}}\\
            \texttt{\#\# Answer}\\
            \texttt{\{answer\}}\\
            \texttt{\#\# Inner thinking}\\
            \bottomrule
        \end{tabular}
    \end{adjustbox}
    \caption{Prompts used to generate reasoning for NQ and TriviaQA in the training data.}
    \label{tab:factual-inner-thinking}
\end{table}

\begin{table}[ht]
    \centering
    \footnotesize
    \begin{adjustbox}{max width=\textwidth}
        \begin{tabular}{p{16cm}}
            \toprule
            \texttt{You will be given a factual question and its short answer. We will also provide you some inner reasoning that leads to the answer.}\\
            \texttt{It is possible that there are multiple answers to the question, and we are only providing one for simplicity.}\\
            \texttt{Your job is to rewrite the short answer such that it is appropriate for spoken form.}\\
            \texttt{When asked a factual question, humans will think about the question in their mind and come up with an answer that summarizes their inner reasoning in a concise and clear response that answers that question and keeps only the relevant information.}\\
            \texttt{Your job is to form this spoken answer based on the question, the inner reasoning, and the short answer.}\\
            \\
            \texttt{\#\# Example 1}\\
            \texttt{\#\# Question:}\\
            \texttt{What color was John Wiles Booth's hair?}\\
            \texttt{\#\# Inner thinking}\\
            \texttt{I think the user is asking about John Wilkes Booth who assassinated United States president Lincoln on April 14, 1865. His hair color is jet‑black. Got the answer and ready to provide it to the user.}\\
            \texttt{\#\# Short answer}\\
            \texttt{jet‑black}\\
            \texttt{\#\# Spoken answer}\\
            \texttt{You mean the one who killed Lincoln? His hair color was jet‑black.}\\
            \\
            \texttt{\#\# Example 2}\\
            \texttt{\#\# Question:}\\
            \texttt{Can you make and receive calls in airplane mode?}\\
            \texttt{\#\# Inner thinking}\\
            \texttt{Airplane mode is a setting available on smartphones and other portable devices. When activated, it suspends all radio‑frequency transmissions, disabling voice and data services. So you can't make or receive calls while it's on.}\\
            \texttt{\#\# Short answer}\\
            \texttt{no}\\
            \texttt{\#\# Spoken answer}\\
            \texttt{Since airplane mode disables all signal transmissions, you cannot make or receive calls in airplane mode.}\\
            \\
            \texttt{\#\# Example 3}\\
            \texttt{\#\# Question:}\\
            \texttt{Who played "Will Munson" on "As the World Turns"?}\\
            \texttt{\#\# Inner thinking}\\
            \texttt{The user likely means the CBS soap opera "As the World Turns." The character Will Munson was most recently portrayed by Jesse Lee Soffer.}\\
            \texttt{\#\# Short answer}\\
            \texttt{Jesse Lee Soffer}\\
            \texttt{\#\# Spoken answer}\\
            \texttt{It's Jesse Lee Soffer—he played Will Munson on that soap opera.}\\
            \\
            \texttt{\# Now it is your turn}\\
            \texttt{\#\# Question:}\\
            \texttt{\{question\}}\\
            \texttt{\#\# Inner thinking}\\
            \texttt{\{reasoning\}}\\
            \texttt{\#\# Short answer}\\
            \texttt{\{answer\}}\\
            \texttt{\#\# Spoken answer}\\
            \bottomrule
        \end{tabular}
    \end{adjustbox}
    \caption{Prompts used to rewrite the original answer into spoken forms for NQ and TriviaQA.}
    \label{tab:spoken-answer-factual}
\end{table}

\begin{table}[ht]
    \centering
    \footnotesize
    \begin{adjustbox}{max width=\textwidth}
        \begin{tabular}{p{16cm}}
            \toprule
            \texttt{\# Task Overview: Inner Monologue Creation}\\
            \texttt{A voice assistant receives a user query or instruction and produces an answer. We now want to come up with a brief “inner monologue” that the assistant might have before speaking.}\\
            \texttt{User Question/Instruction: \{question\}}\\
            \texttt{Assistant Answer: \{answer\}}\\
            \\
            \texttt{\# Task Introduction}\\
            \texttt{Generate a concise inner thinking that the assistant could have had before speaking. Your inner monologue should include understanding of the user’s question or instruction and then a draft answer.}\\
            \texttt{Make sure the inner monologue will lead to the assistant’s provided answer.}\\
            \\
            \texttt{\# Examples}\\
            \\
            \texttt{\#\# Example 1}\\
            \texttt{User Question/Instruction: How can I stop being afraid of the dark?}\\
            \texttt{Assistant Answer: It’s completely normal to feel afraid of the dark, and there are steps you can take to feel more comfortable. Start by gradually exposing yourself to the dark in small, manageable steps. You can use a nightlight as a transitional tool. Establish a bedtime routine that makes you feel relaxed, like reading a book or listening to calming music. Breathing exercises or meditation can help reduce anxiety. Rearrange your bedroom to make it feel safer and more inviting. Remind yourself that the fear is often about not knowing what’s there, rather than actual dangers. Talking to someone about your fears can also be really helpful.}\\
            \texttt{Inner Monologue: The user wants to stop being afraid of the dark. I will first acknowledge that it’s normal to feel fear, then suggest gradual exposure, a comforting bedtime routine, anxiety‑reducing exercises, and talking to someone.}\\
            \\
            \texttt{\#\# Example 2}\\
            \texttt{User Question/Instruction: Can you count down from 100 by subtracting 7 each time?}\\
            \texttt{Assistant Answer: Yes, I can. Help you with counting down from 100 and minus 7 each time. Here we go: 100, 93, 86, 79, 72, 65, 58. Do you need more?}\\
            \texttt{Inner Monologue: The user wants a countdown by sevens from 100. I'll compute 100-7=93, then subtract 7 each step until I have a sequence. I'm ready to present the numbers.}\\
            \\
            \texttt{\#\# Example 3}\\
            \texttt{User Question/Instruction: What are some interesting things I can mix into my scrambled eggs?}\\
            \texttt{Assistant Answer: Scrambled eggs can be a delightful canvas for a variety of ingredients! You can try adding cheese like cheddar, feta, or Swiss for extra creaminess. Fresh herbs such as chives, parsley, or dill can bring a burst of flavor. Diced vegetables like bell peppers, tomatoes, spinach, or mushrooms can add texture and nutrition. For protein, consider mixing in smoked salmon, ham, or crumbled bacon. If you enjoy a bit of heat, a dash of hot sauce or chopped jalapeños can be great. Even a bit of sour cream or cream cheese can make the eggs extra creamy. Enjoy experimenting to find your perfect blend!}\\
            \texttt{Inner Monologue: The user is asking for mix‑ins for scrambled eggs. I’ll recall common add‑ins like cheeses, herbs, vegetables, proteins, and creams, then frame them as suggestions.}\\
            \\
            \texttt{\# Your Task}\\
            \texttt{Generate the inner monologue for this instruction/answer pair.}\\
            \texttt{User Question/Instruction: \{question\}}\\
            \texttt{Assistant Answer: \{answer\}}\\
            \texttt{Inner Monologue: (provide the concise inner monologue only)}\\
            \bottomrule
        \end{tabular}
    \end{adjustbox}
    \caption{The prompt used to generate the reasoning for VoiceAssistant400K.}
    \label{tab:voice assistant 400K prompt}
\end{table}

\end{document}